\documentclass[final]{cvpr}

\usepackage{times}
\usepackage{epsfig}
\usepackage{graphicx}
\usepackage{amsmath}
\usepackage{amssymb}

\usepackage[utf8]{inputenc} 
\usepackage[T1]{fontenc}    

\usepackage{url}            
\usepackage{booktabs}       
\usepackage{amsfonts}       
\usepackage{nicefrac}       
\usepackage{microtype}      
\usepackage{times}
\usepackage{epsfig}
\usepackage{graphicx}
\usepackage{amsmath}
\usepackage{float}
\usepackage{gensymb}
\usepackage{amssymb}
\usepackage{subcaption}
\usepackage[noline,ruled]{algorithm2e}
\usepackage[titletoc]{appendix}
\usepackage{cuted}
\usepackage{capt-of}

\usepackage[pagebackref=true,breaklinks=true,colorlinks,bookmarks=false]{hyperref}

\def\cvprPaperID{10684} 
\def\confYear{CVPR 2021}

\begin{document}


\author{Adel Ahmadyan, Liangkai Zhang, Jianing Wei, Artsiom Ablavatski, Matthias Grundmann\\
{\tt\small \{ahmadyan, liangkai, jianingwei, artsiom, grundman\}@google.com}
}

\title{Objectron: A Large Scale Dataset of Object-Centric Videos in the Wild with Pose Annotations}

\maketitle

\begin{strip}\centering
 \includegraphics[width=0.15\textwidth]{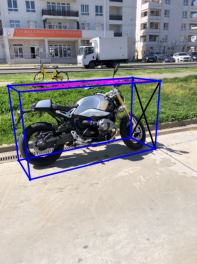}
    \includegraphics[width=0.15\textwidth]{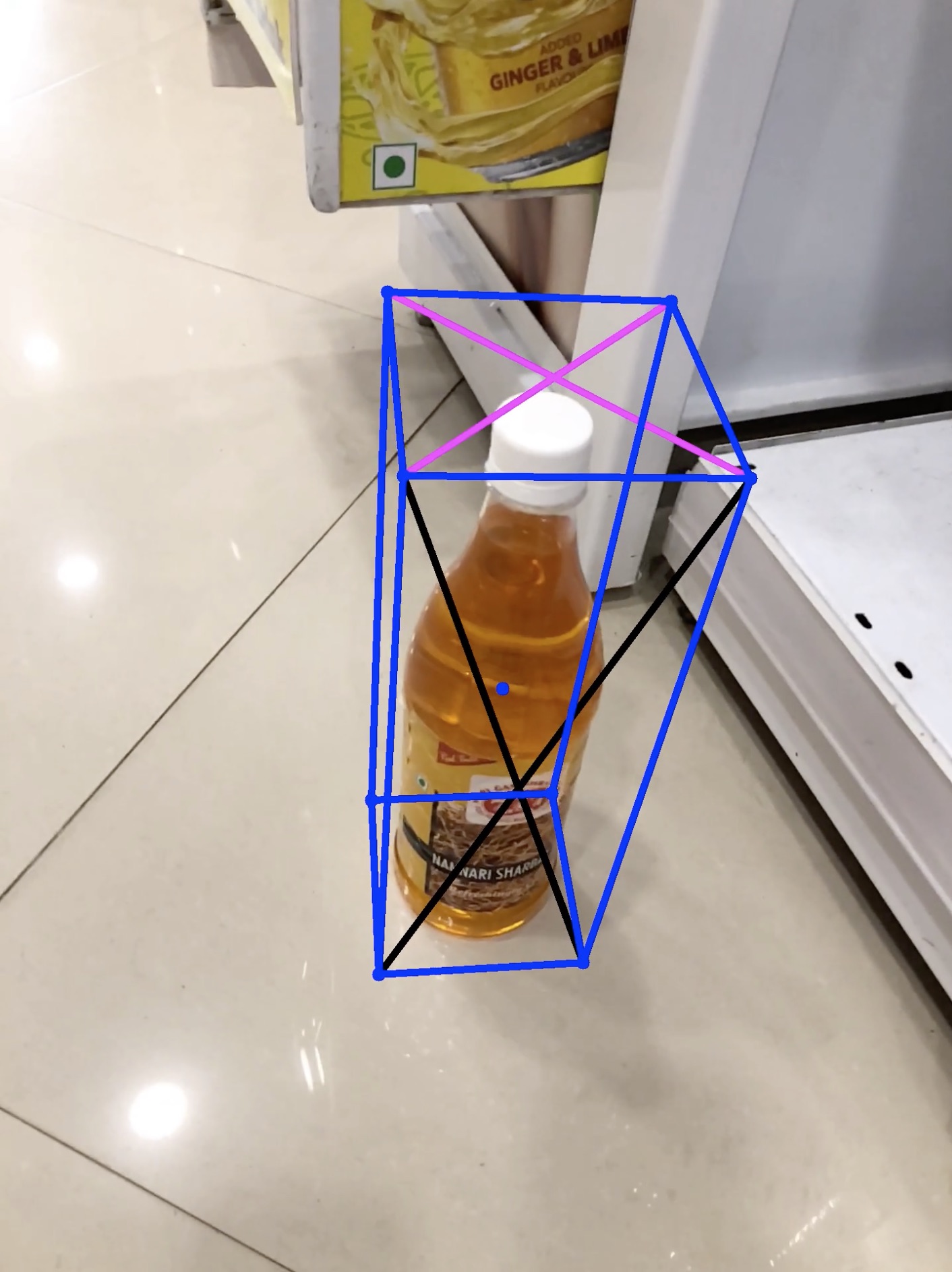}
    \includegraphics[width=0.15\textwidth]{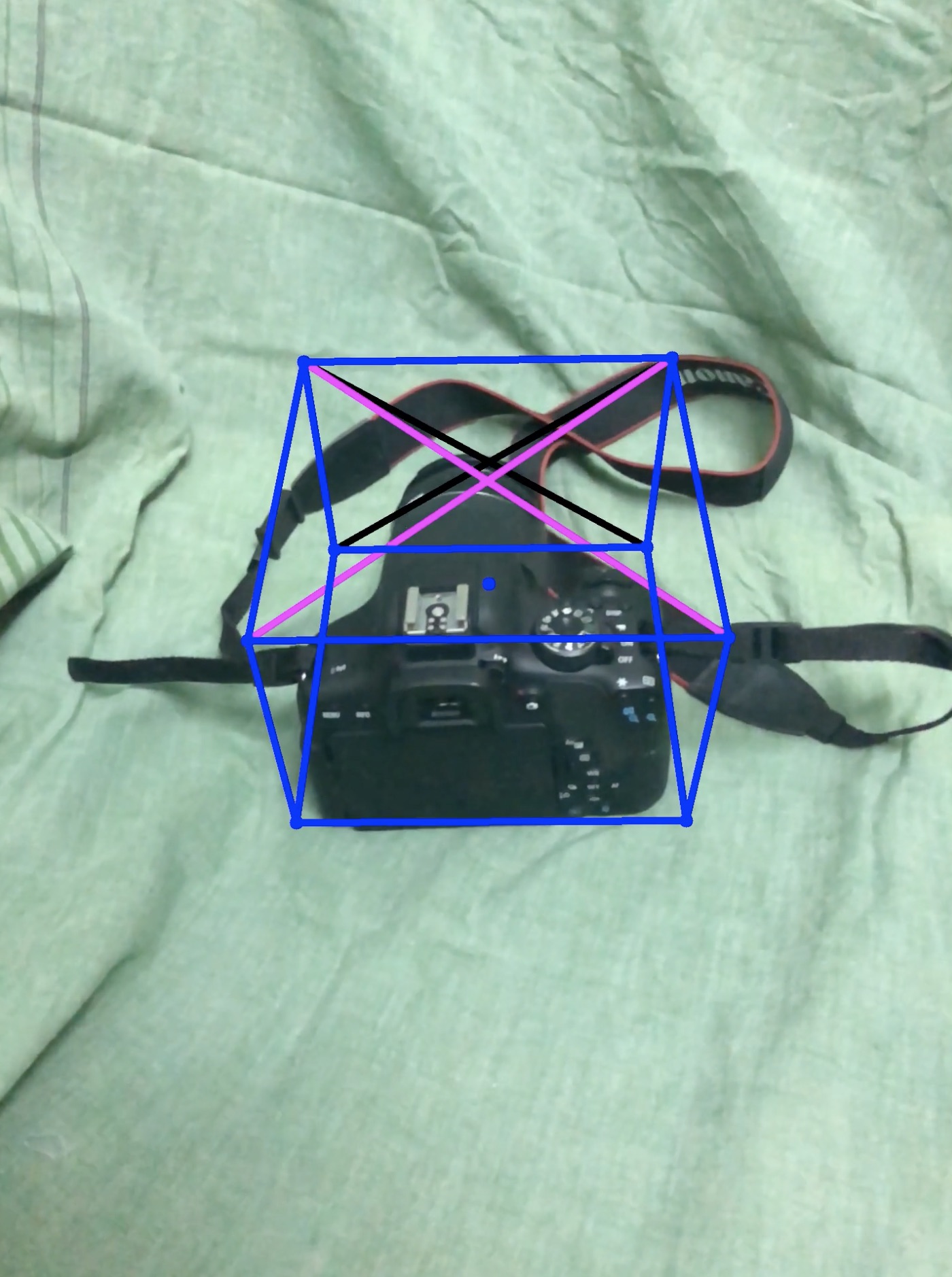}
    \includegraphics[width=0.15\textwidth]{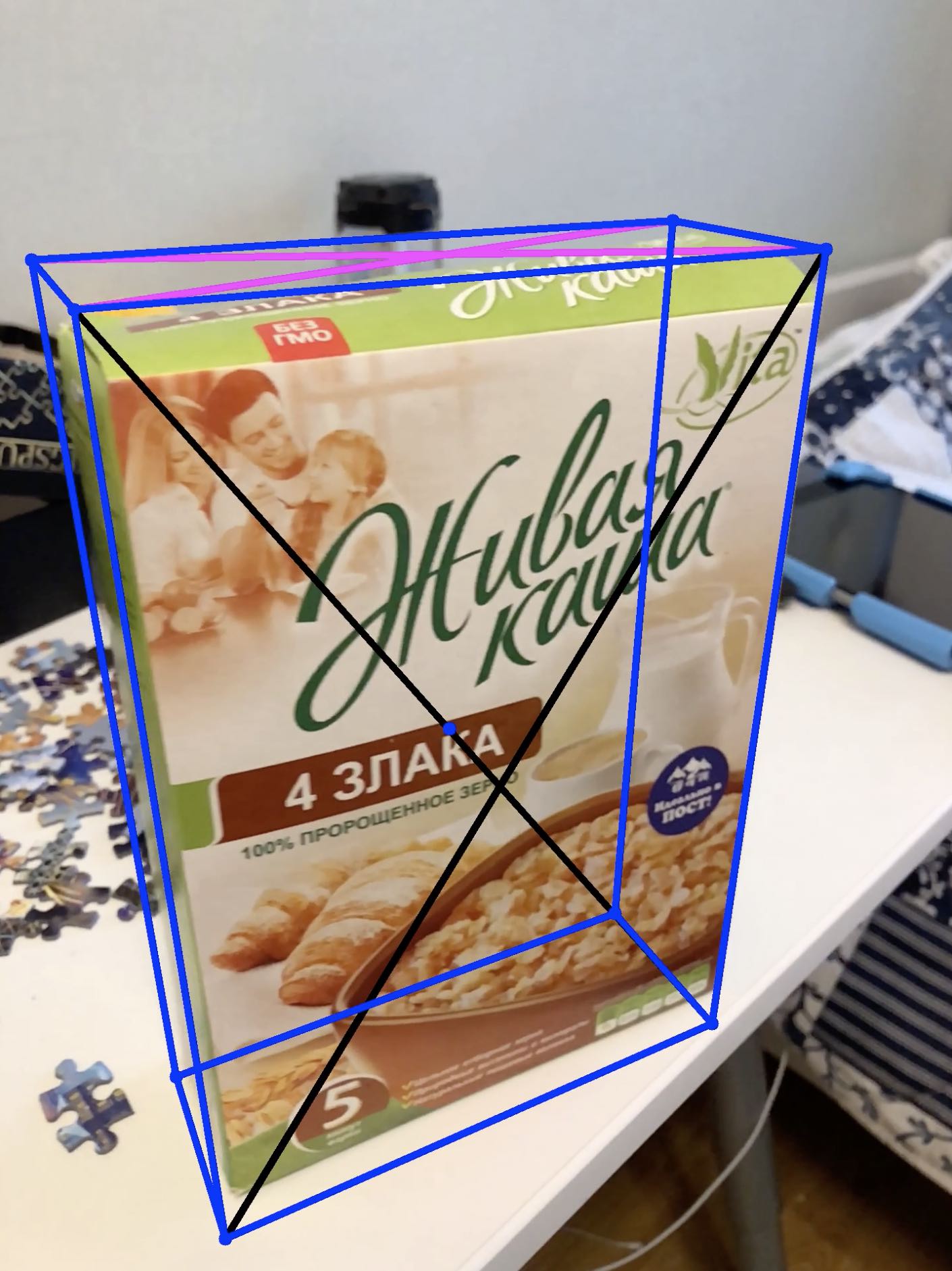}
    \includegraphics[width=0.15\textwidth]{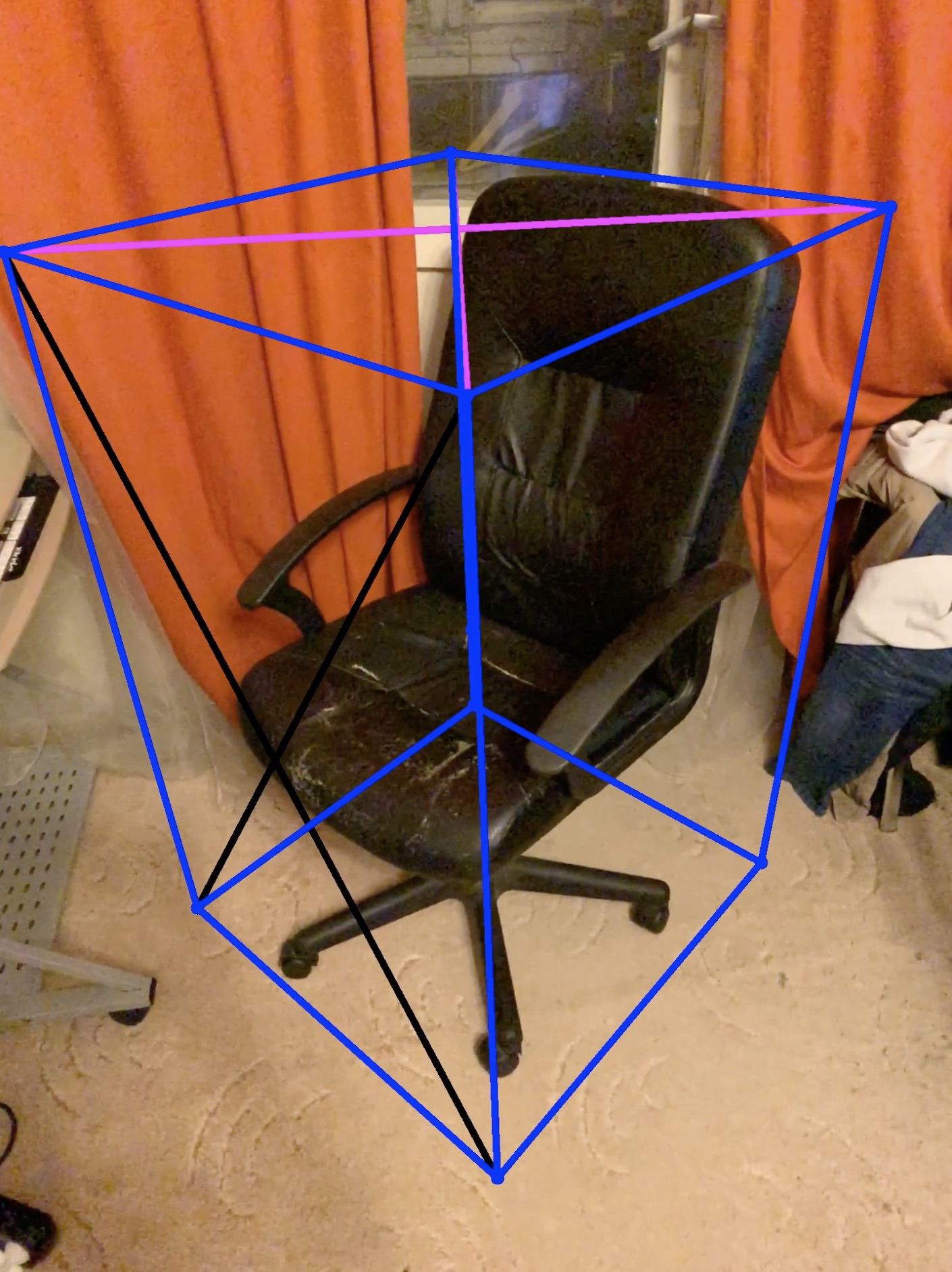}

    \includegraphics[width=0.15\textwidth]{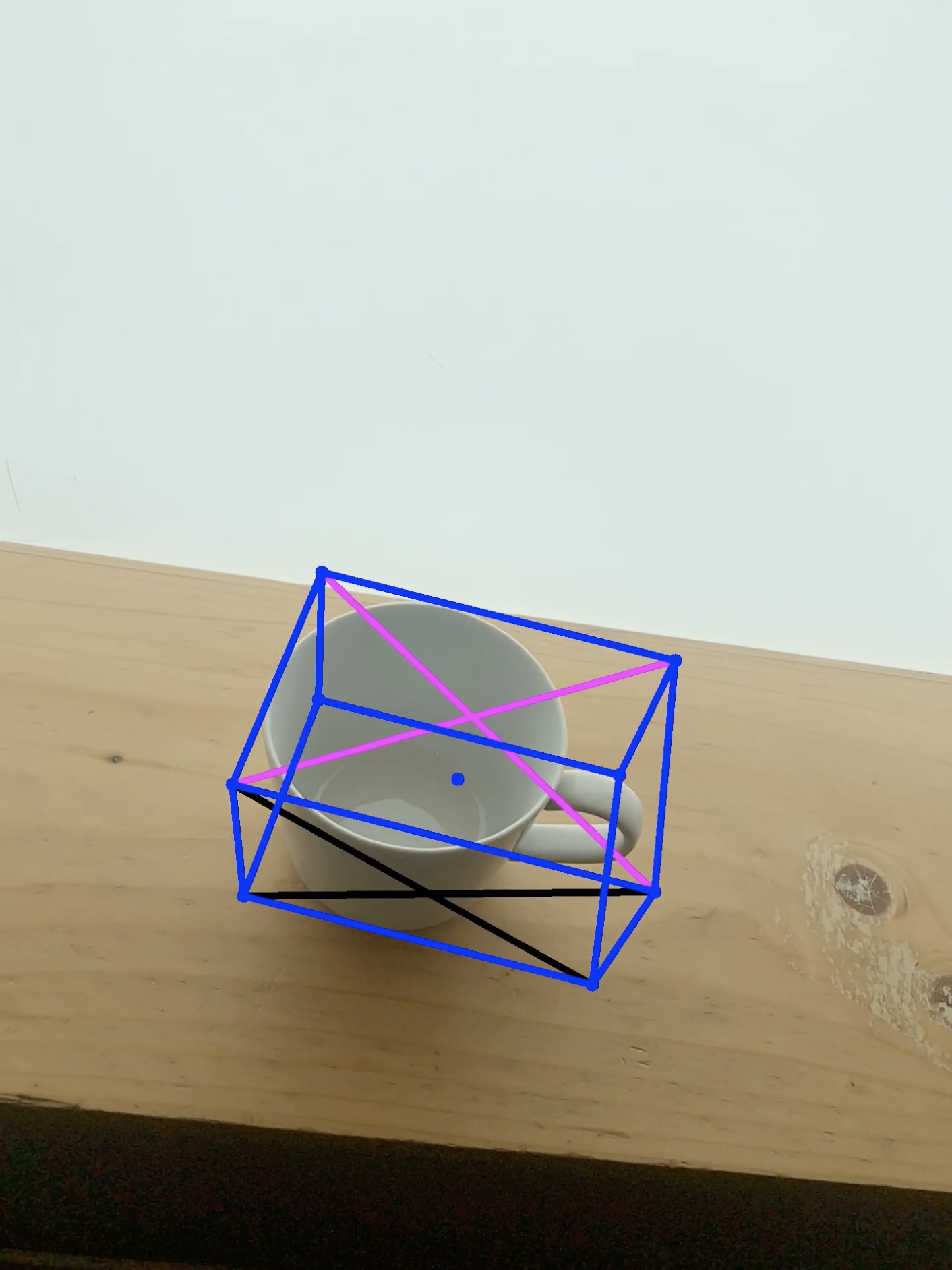}
    \includegraphics[width=0.15\textwidth]{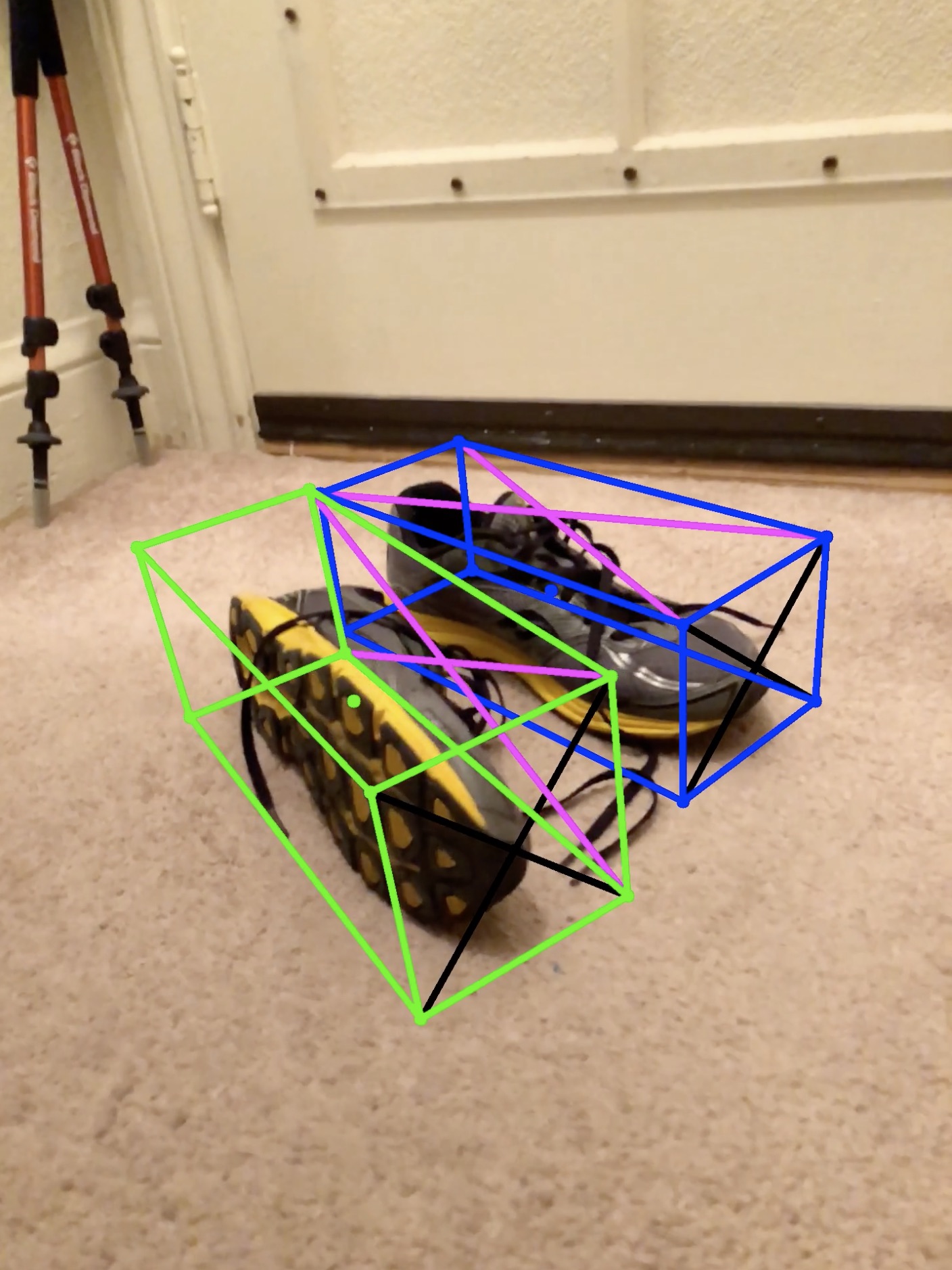}
    \includegraphics[width=0.15\textwidth]{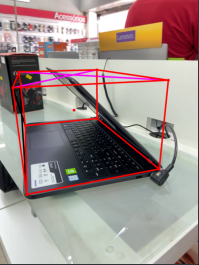}
    \includegraphics[width=0.15\textwidth]{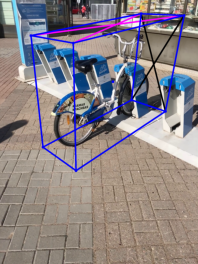}
    \includegraphics[width=0.15\textwidth]{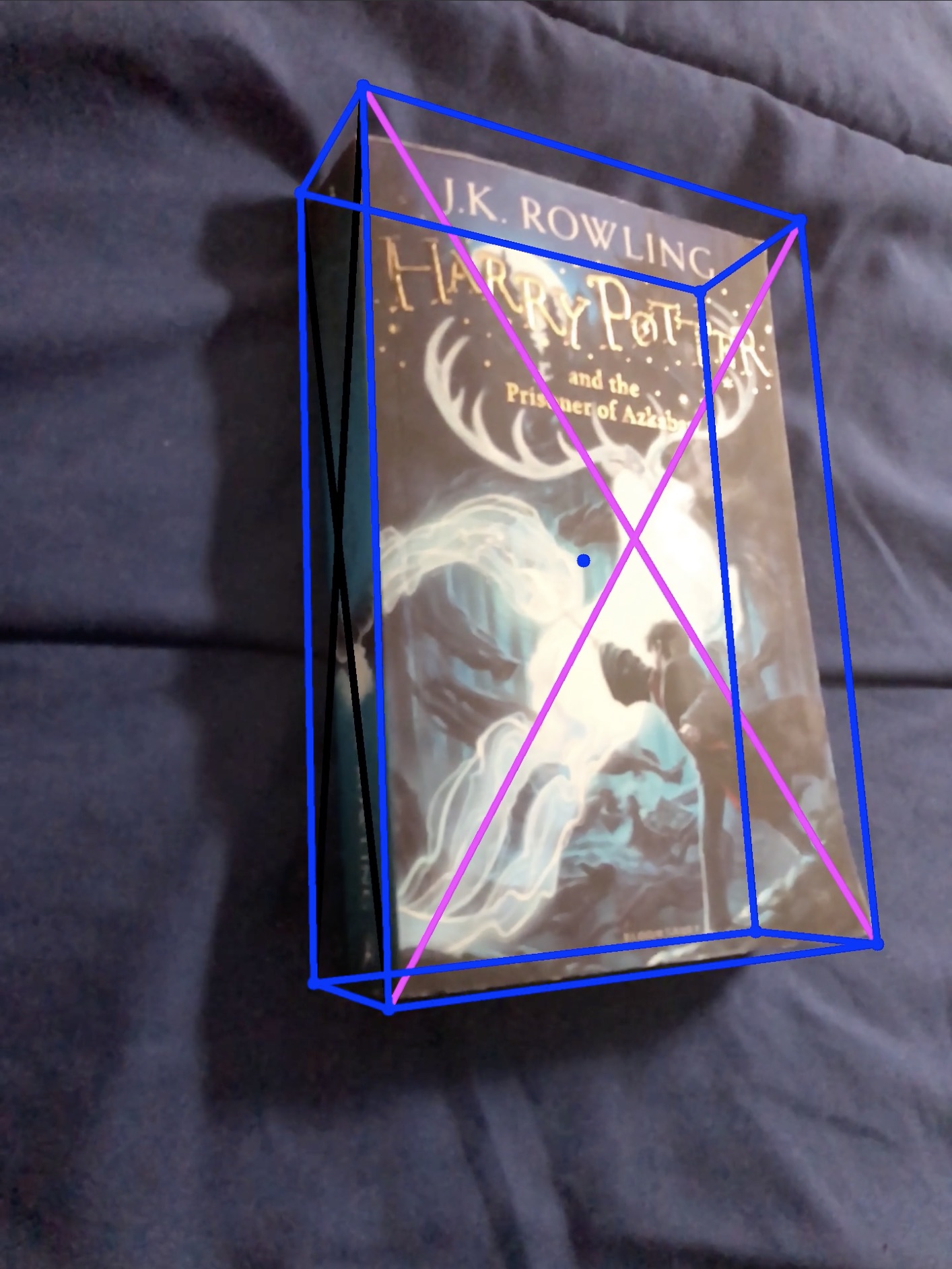}
    
\captionof{figure}{Example videos in our dataset. Each sample is a video annotated with the object's 3D bounding box.
\label{fig:feature-graphic}}
\end{strip}

\begin{abstract}
   3D object detection has recently become popular due to many applications in robotics, augmented reality, autonomy, and image retrieval. 
   We introduce the Objectron dataset to advance the state of the art in 3D object detection and foster new research and applications, such as 3D object tracking, view synthesis, and improved 3D shape representation. 
   The dataset contains object-centric short videos with pose annotations for nine categories and includes 4 million annotated images in $14,819$ annotated videos. 
   We also propose a new evaluation metric, 3D Intersection over Union, for 3D object detection. We demonstrate the usefulness of our dataset in 3D object detection tasks by providing baseline models trained on this dataset.
   Our dataset and evaluation source code are available online at \url{http://www.objectron.dev}.
\end{abstract}
\section{Introduction} \label{sec:intro}

\begin{figure*}[tb]
    \centering
    \includegraphics[width=\textwidth]{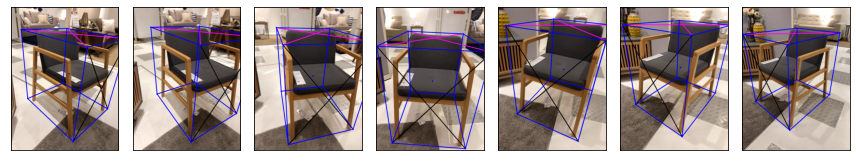}
    \caption{Our dataset consists of object-centric videos, which capture different views of the same objects from different angles.}
    \label{fig:example-seq}
\end{figure*}

The state of the art in machine learning has achieved exceptional accuracy on many computer vision tasks solely by training models on images. Building upon these successes and advancing 3D object understanding has great potential to power a wider range of applications, such as augmented reality, robotics, autonomy, and image retrieval. Yet, understanding objects in 3D remains a challenging task due to the lack of large real-world datasets compared to 2D tasks (e.g., ImageNet~\cite{Deng:2009jn}, COCO~\cite{Lin:2014vm}, and Open Images~\cite{Kuznetsova:2020ff}). To empower the research community for continued advancement in 3D object understanding, there is a strong need for the release of object-centric video datasets, which capture more of the 3D structure of an object, while matching the data format used for many down-stream vision tasks (i.e., video or camera streams), to aid in the training and benchmarking of machine learning models.
The object-centric approach is consistent with how our brains perceive new objects too. For example, when a child wants to learn the shape of a chair, they'll walk around and look at the chair from different angles to pick up information. In other words, "We must also move in order to perceive"~\cite{Gibson:1979}\cite{Zamir:2018}.

We present the Objectron dataset, a collection of short, object-centric video clips capturing a larger set of common objects from different angles. Each video clip is accompanied by augmented reality (AR) session metadata that includes camera poses, sparse point-clouds, and surface planes. The data also contains manually annotated 3D bounding boxes for each object, which describe the object’s position, orientation, and dimensions. The dataset consists of $14,819$ annotated video clips and $4M$ annotated images collected from a geo-diverse sample (covering ten countries across five continents). Figure~\ref{fig:example-seq} shows an example video in our dataset.

There are several advantages of using Objectron dataset over existing works:
\begin{itemize}
    \item Videos contain multiple views of the same object, enabling many applications well beyond 3D object detection. This includes multi-view geometric understanding, view synthesis, 3D shape reconstruction, etc.
    \item The 3D bounding box is present in the entire video and is temporally consistent, thus enabling 3D tracking applications.
    \item Our dataset is collected in the wild to provide better generalization for real-world scenarios in contrast to datasets that are collected in a controlled environment ~\cite{Hinterstoisser:2012ik}\cite{Calli:2017cd}.
    \item Each instance's translation and size are stored in metric scale, thanks to accurate on-device AR tracking and provides sparse point clouds in 3D, enabling sparse depth estimation techniques. The images are calibrated and the camera parameters are provided, enabling the recovery of the object's true scale.
    \item Our annotations are dense and continuous, unlike some of the previous work~\cite{Savarese:2007by} where viewpoints have been discretized to fit into bins.
    \item Each object category contains hundreds of instances, collected from different locations across different countries in different lighting conditions.
\end{itemize}

We have conducted experiments for 3D object detection tasks using our dataset. We proposed a novel method to compute the precise 3D IoU of oriented 3D bounding boxes. These experiments can be used as baselines for future research. 
\section{Previous Work} \label{sec:prev-work}
In this section, we review several commonly used datasets for 3D object detection and compare them with Objectron. 

BOP challenge~\cite{Hodan:2018ie} consists of a set of benchmark for 3D object detection and combines many of these smaller datasets into a larger one. Most of the images in the dataset are taken in a very controlled environment and feature clutter and heavy occlusion that apppear in industrialized setups. T-LESS~\cite{Hodan:2017ec} features industrialized objects that lack texture or color. Rutgers APC~\cite{Rennie:2016gm} contains 14 textured objects from the Amazon picking challenge. LineMOD~\cite{Hinterstoisser:2012ik} is the most commonly used dataset for object pose estimation. Similar to LineMOD, IC-BIN dataset~\cite{Doumanoglou:2016} adds a few more categories. YCB~\cite{Calli:2017cd} dataset contains videos of objects and their poses in a controlled environment. In comparison to these datasets, Objectron has a larger scale and contains high-resolution videos of common objects in the wild.

ObjectNet3D~\cite{Xiang:2016} is dataset that contains 3D object poses from images. Similarly, Pascal3D+~\cite{Xiang:14} adds 3D pose annotations to the Pascal VOC and a few images from the ImageNet dataset. In comparison to Objectron, these datasets contain more categories and instances but they only include images. Furthermore, the objects are annotated with a 2D-to-3D alignment process, so their pose is annotated up to 6-DoF (instead of nine) and camera intrinsics are not available. So it is not possible to recover the object's scale from these datasets. Pix3D~\cite{Sun:2018uq} contains pixel-level 2D-3D pose alignment. Similarly, 3DObject\cite{Savarese:2007by} provides discretized viewpoint annotations for 10 everyday object categories.

Another type of dataset used for 3D object detection tasks is scene datasets. In these datasets, a video of a scene is recorded with an RGBD camera or LIDAR. ScanNet~\cite{Dai:2017} is a large scale video dataset of indoor scenes with semantic annotations. The dataset contains over 1500 scenes recorded in RGBD videos. The dataset does not contain 3D pose information, however, Scan2CAD~\cite{Avetisyan:2018ts} annotates the original scans with ShapeNetCore~\cite{Chang:2015uq} models to label each object's pose. Rio~\cite{Wald:2019te} is another dataset that contains indoor scans annotated with an object's 3D pose. In comparison Objectron videos are object-centric, and we have an order of magnitude more samples in our dataset. 

Another approach is to use synthetic data. ShapeNet~\cite{Chang:2015uq} includes CAD models for many objects and has been widely used~\cite{Tulsiani:2020wv}\cite{Sridhar:2019vr}, However, the sample's visual quality is limited and it does not generalize well to real-world applications. Synthetic datasets offer valuable data for training and benchmarking, but the ability to generalize to the real-world is unknown. HyperSim\cite{Roberts:2020} generates photo-realistic scenes with object pose annotation, but reproducibility is challenging since they have not released any of the dataset's assets.


\section{Data Collection and Annotation} \label{sec:data}

\subsection{Object Categories} \label{sec:categories}
In Objectron dataset, the aim was to select meaningful categories of common objects that form a representative set of all categories that are practically relevant and technically challenging. Our goal was to capture these objects in their common environment, and in relative context whether it would be in store, indoor or outdoor environment. We also included objects of various sizes, ranging from a few centimeters (e.g. cups) to as large as chairs and bikes.

The object categories in the dataset contain both rigid, and non-rigid objects. We included non-rigid categories such as bikes and laptops specifically since we expect techniques using CAD models or strong priors will face challenges estimating the pose of these object categories. We should mention non-rigid objects remain stationary during the period of each video.

Many 3D object detection models are known to exhibit difficulties in estimating rotations of symmetric objects~\cite{Labbe:2020vd}. 
Symmetric objects have ambiguity in their one, two, or even three degrees of rotation. Therefore we added categories like "cups" and "bottles" specifically to test it.

It has been shown that vision models pay special attention to texts in the images \cite{Geirhos:2018vs}. Re-producing texts and labels correctly are important in generative models too. Therefore we added categories of objects with very distinct texts in their labels such as "books" and "cereal boxes". Since our dataset is collected from a geo-diverse set of countries, multiple different languages are present in the videos as well. We should report these categories, despite having relatively simple box-shaped geometry, have very different texture patterns. So our baseline experiments have difficulty estimating their poses accurately.

Since we strive for real-time perception we included a few categories (shoes and chairs) that enable exciting applications, such as augmented reality and image retrieval.

\subsection{Data Collection} \label{sec:collection}
Our data collection consists of the video recording when the camera moves around the object and looks at it from different angles. We also capture camera poses, point-clouds, and surface planes through an AR session (e.g. ARKit~\cite{arkit} or ARCore~\cite{arcore}). AR solutions track a set of features through the video and estimate their 3D coordinates on-device in real-time. Meanwhile, they estimate camera poses using bundle-adjustment and filtering. We emphasize all the translation and scale reported in our dataset are on the metric scale. We released both of these files (the video recording and the AR session metadata) in the dataset. We assume the standard pinhole camera model, and provide calibration, extrinsics and intrinsics matrix for every frame in the dataset.

All the videos are recorded in $1440\times1920$ resolution at $30fps$ using the back-camera of high-end phones. We only enabled data collection on a very few phone models (<5) to make sure the output quality remains consistent across our dataset. To avoid drift in the AR session, we kept the video length short at around $10sec$. The collectors are instructed to avoid rapid movements to avoid blurry images. Finally, the object remains stationary in all of our videos.

\begin{figure}[tb]
    \centering
    \includegraphics[width=\columnwidth]{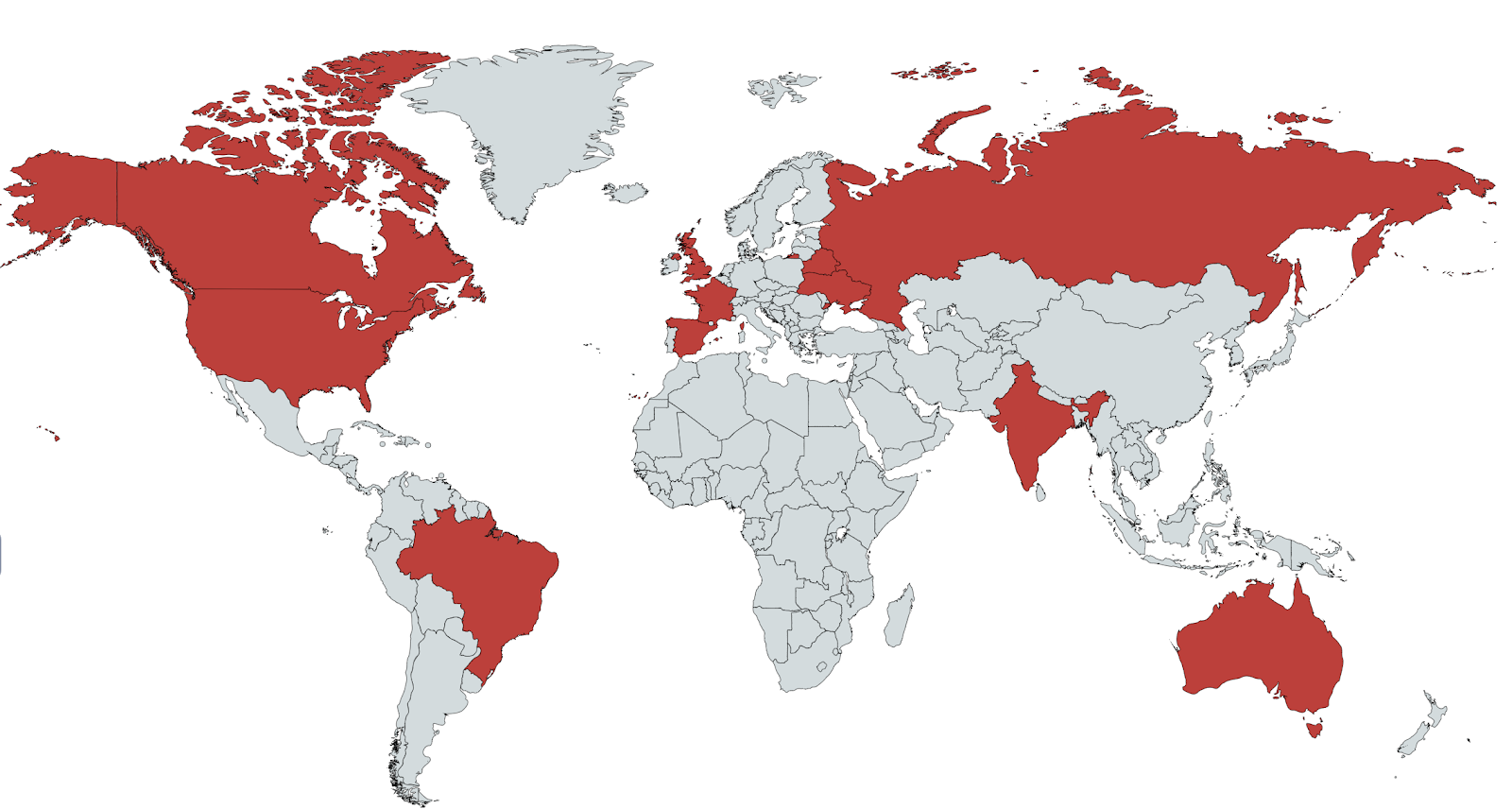}
        \caption{Countries where we collected data from.}
        \label{fig:map}
\end{figure}

By using mobile phones for data collection we were able to quickly launch a data collection campaign across many different countries. The dataset is geo-diverse and covers 10 countries over five continents. Figure~\ref{fig:map} shows which countries are represented in the dataset. The samples are distributed uniformly in each region.

\subsection{Data Annotation} \label{sec:annotation}
\begin{figure}[H]
    \centering
    \begin{subfigure}{\columnwidth}
        \centering
        \includegraphics[width=0.9\columnwidth]{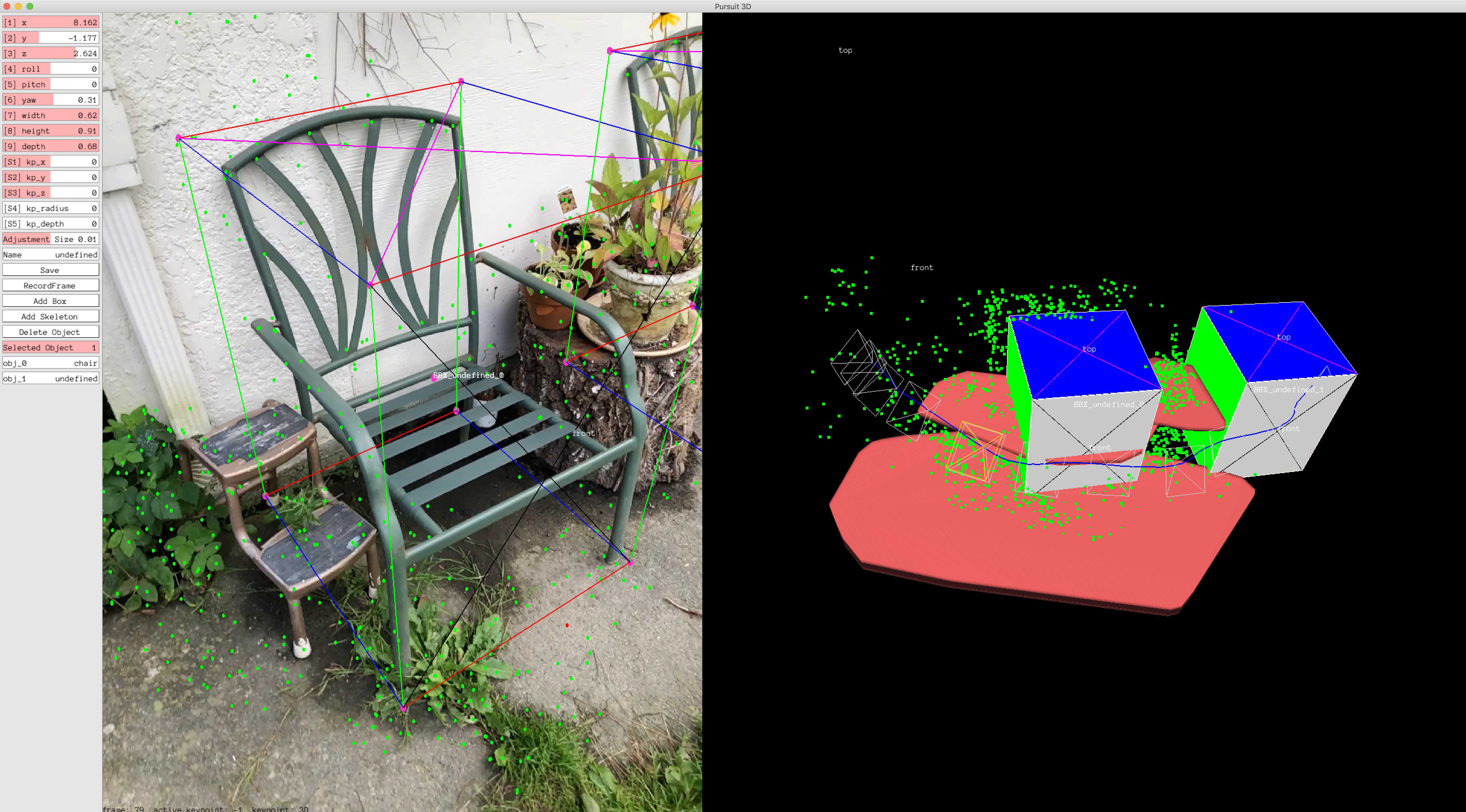}
        \caption{Our annotation tool}
        \label{fig:annotation_tool}
    \end{subfigure}
    \begin{subfigure}[t]{0.3\columnwidth}
        \includegraphics[width=\textwidth]{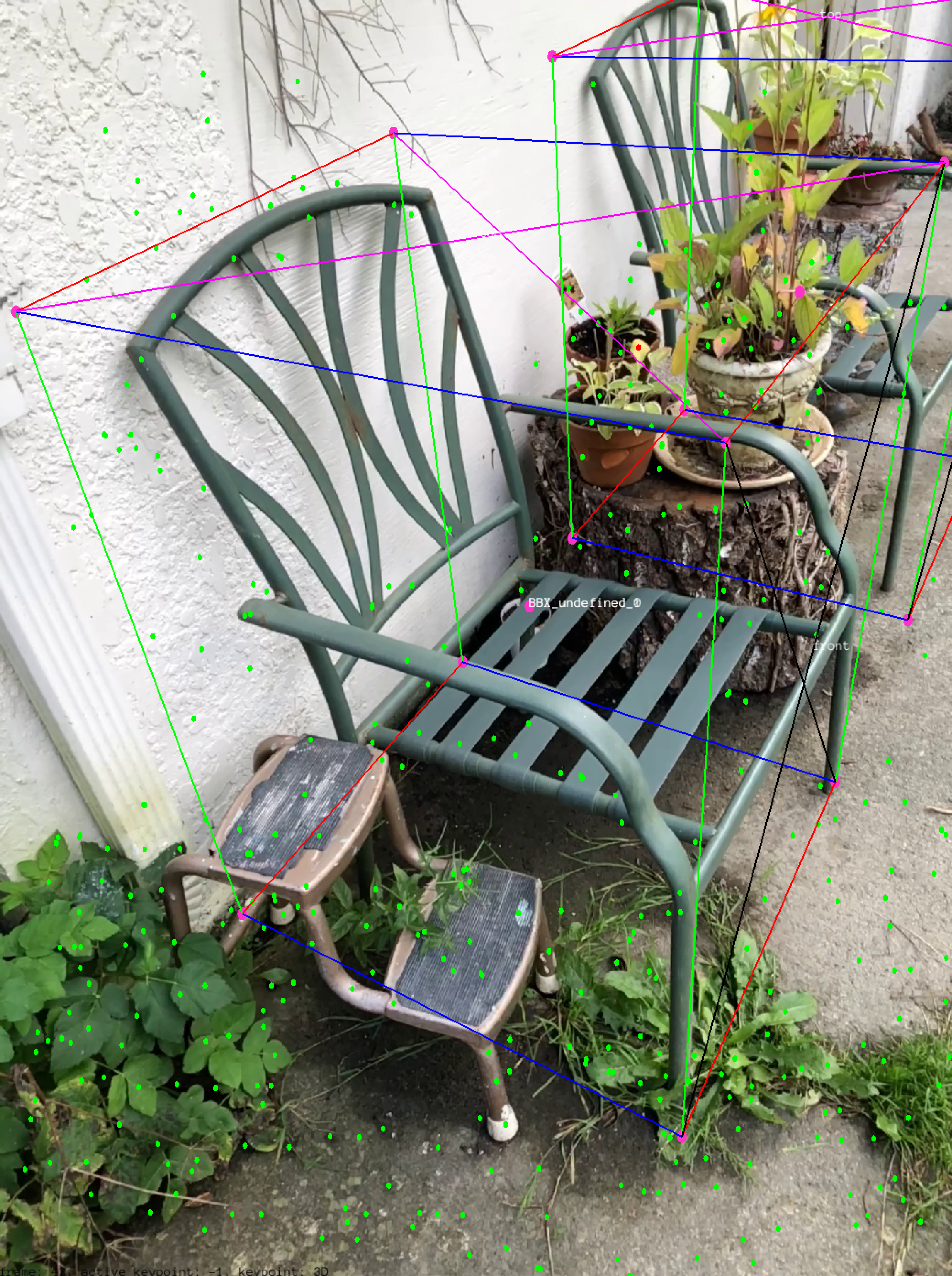}
        \label{fig:view-1}
    \end{subfigure}
    \begin{subfigure}[t]{0.3\columnwidth}
        \includegraphics[width=\textwidth]{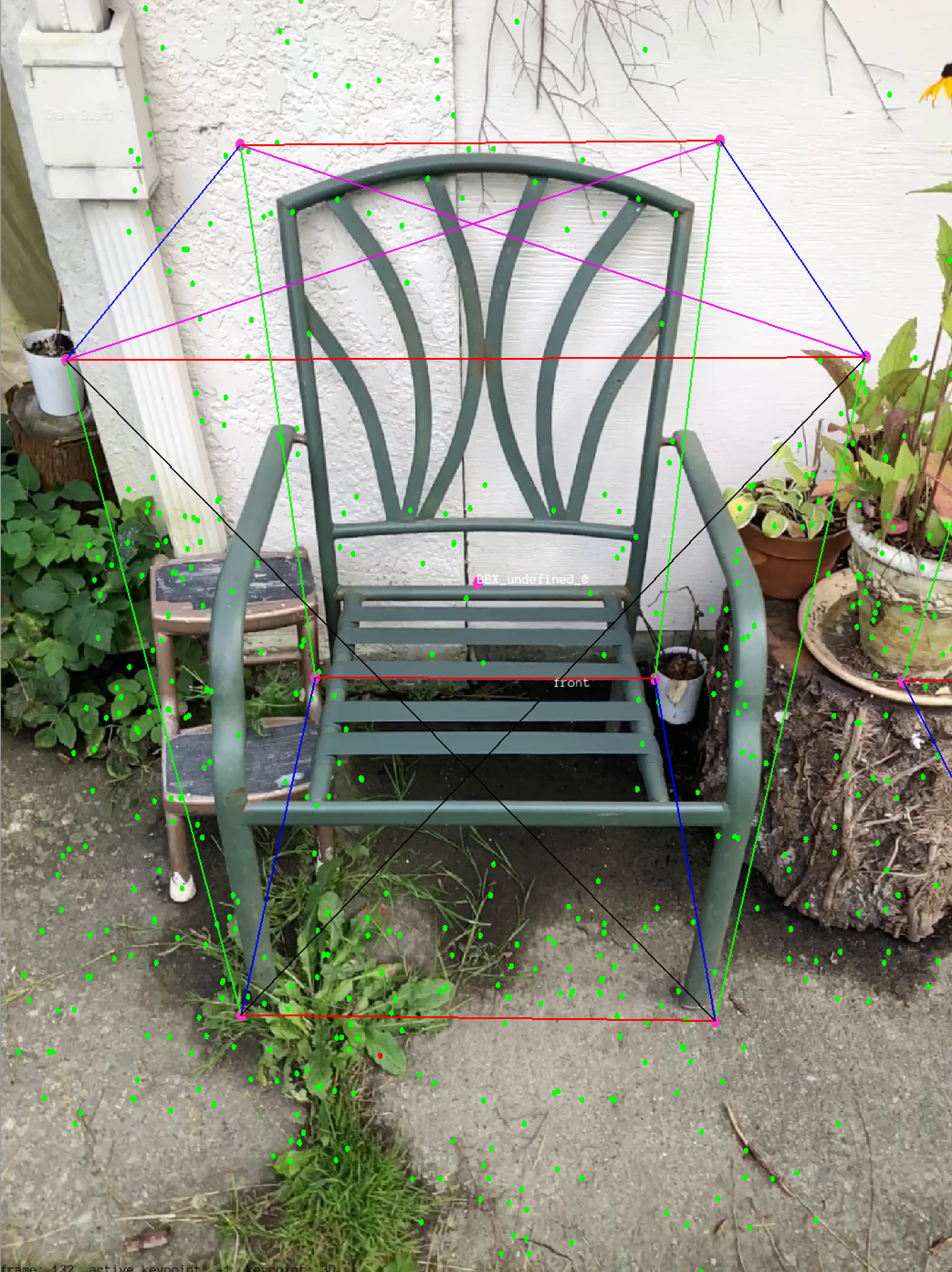}
        \label{fig:view-2}
    \end{subfigure}
    \begin{subfigure}[t]{0.3\columnwidth}
        \includegraphics[width=\textwidth]{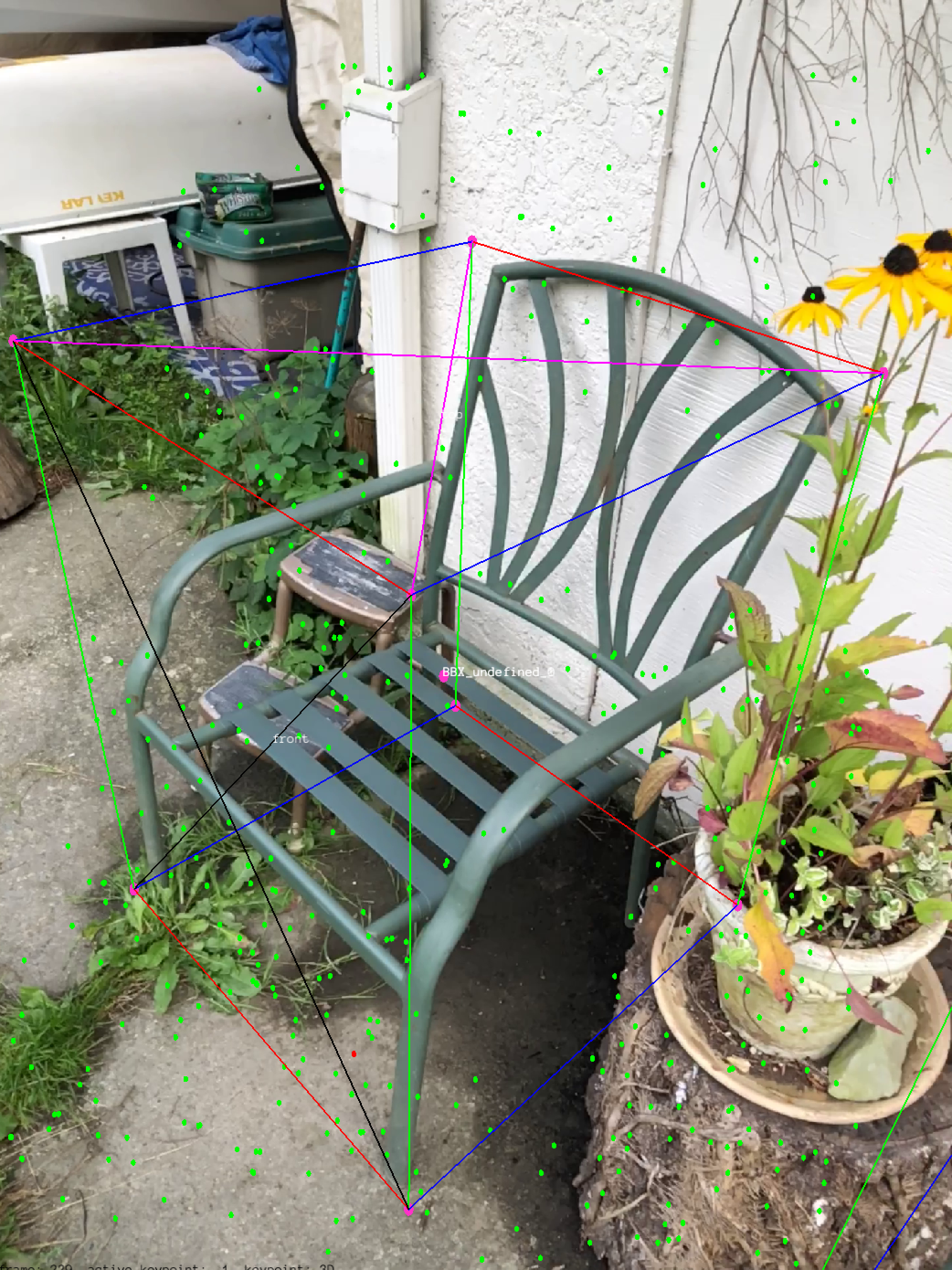}
        \label{fig:view-2}
    \end{subfigure}
    \caption{Data annotation. The annotated 3D box is verified at multiple views, and then populated to all images in the sequence.}
    \label{fig:multi-view}
\end{figure}

Efficient and accurate data annotation is the key to building large-scale datasets. Annotating 3D bounding boxes for each image is time-consuming and expensive. Instead, we annotate 3D objects in a video clip and populate them to all frames in the clip, scaling up the annotation process, and reducing the per image annotation cost. The user interfaces for the annotation tool is shown in Figure~\ref{fig:annotation_tool}. The input to our annotation tool is a video sequence of stationary objects. It covers different viewing angles of objects.

Next, we show the 3D world map to the annotator side-by-side with the images from the video sequence (Figure~\ref{fig:annotation_tool}). The annotator draws a 3D bounding box in the 3D world map, and our tool projects the 3D bounding box over all the frames given pre-computed camera poses from the AR sessions (such as ARKit or ARCore). The annotator looks at the projected bounding box and makes necessary adjustments (position, orientation, and the scale of a 3D bounding box) so the projected bounding box looks consistent across different frames. At the end of the process, the user saves the 3D bounding box and the annotation. 

The benefits of our approach are 1) by annotating a video once, we get annotated images for all frames in the video sequence; 2) by using AR, we can get accurate metric sizes for bounding boxes.

\subsection{Annotation Variance} \label{sec:annotation-variance}
\begin{figure}[t]
    \centering
    \begin{subfigure}{\columnwidth}
        \includegraphics[width=\textwidth]{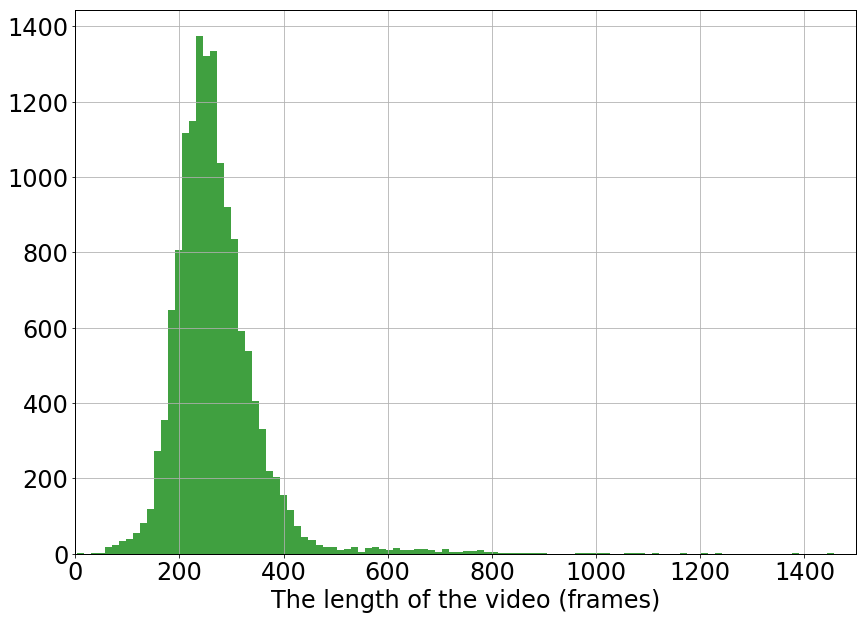}
    \end{subfigure}
    \caption{Distribution of the video length in our dataset. Majority of the videos are 10 seconds long (300 frames) and the longest video is 2022 frames long.}
    \label{fig:length}
\end{figure}

\begin{figure*}[t]
    \centering
    \includegraphics[width=0.15\textwidth]{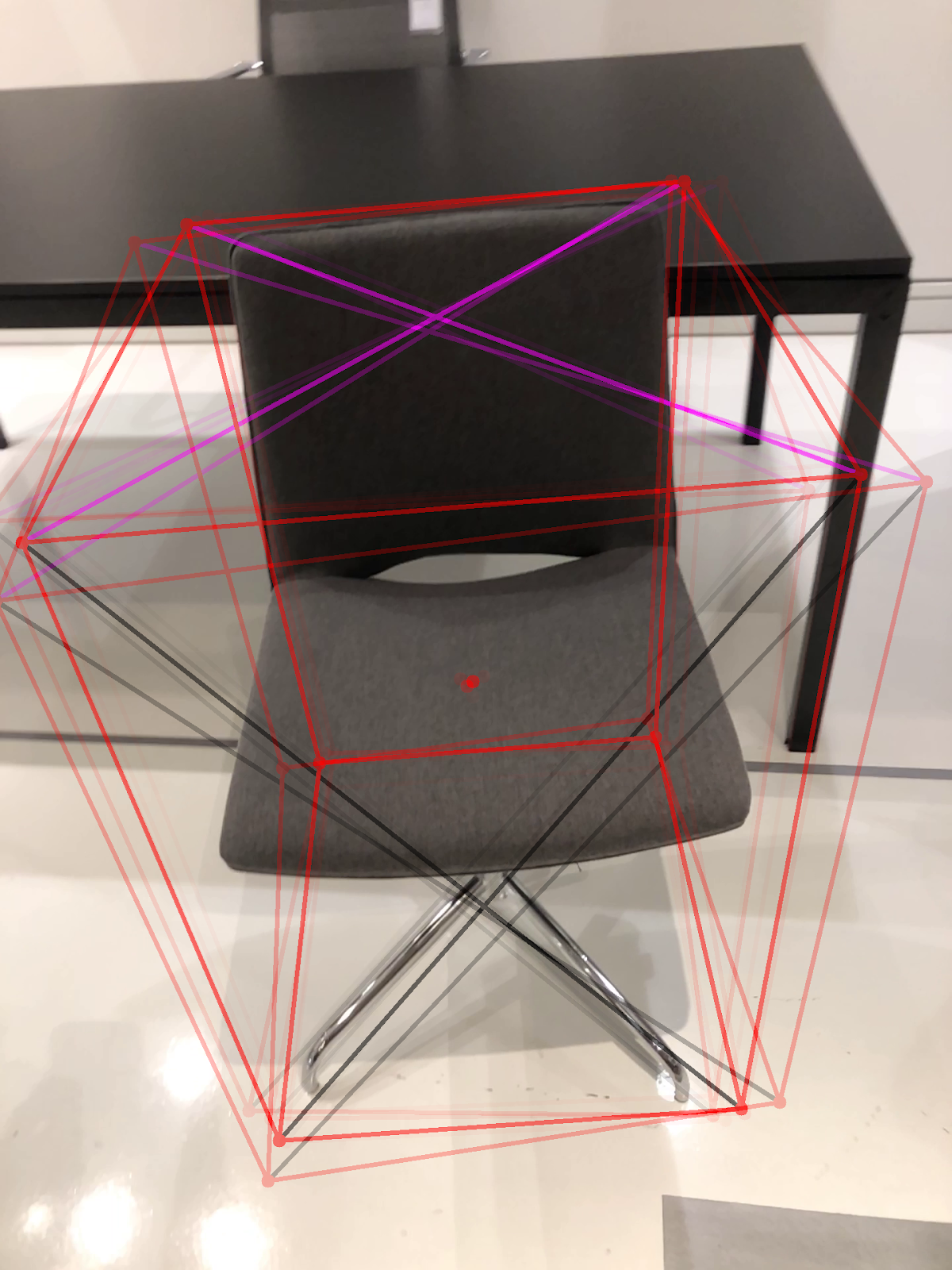}
    \includegraphics[width=0.15\textwidth]{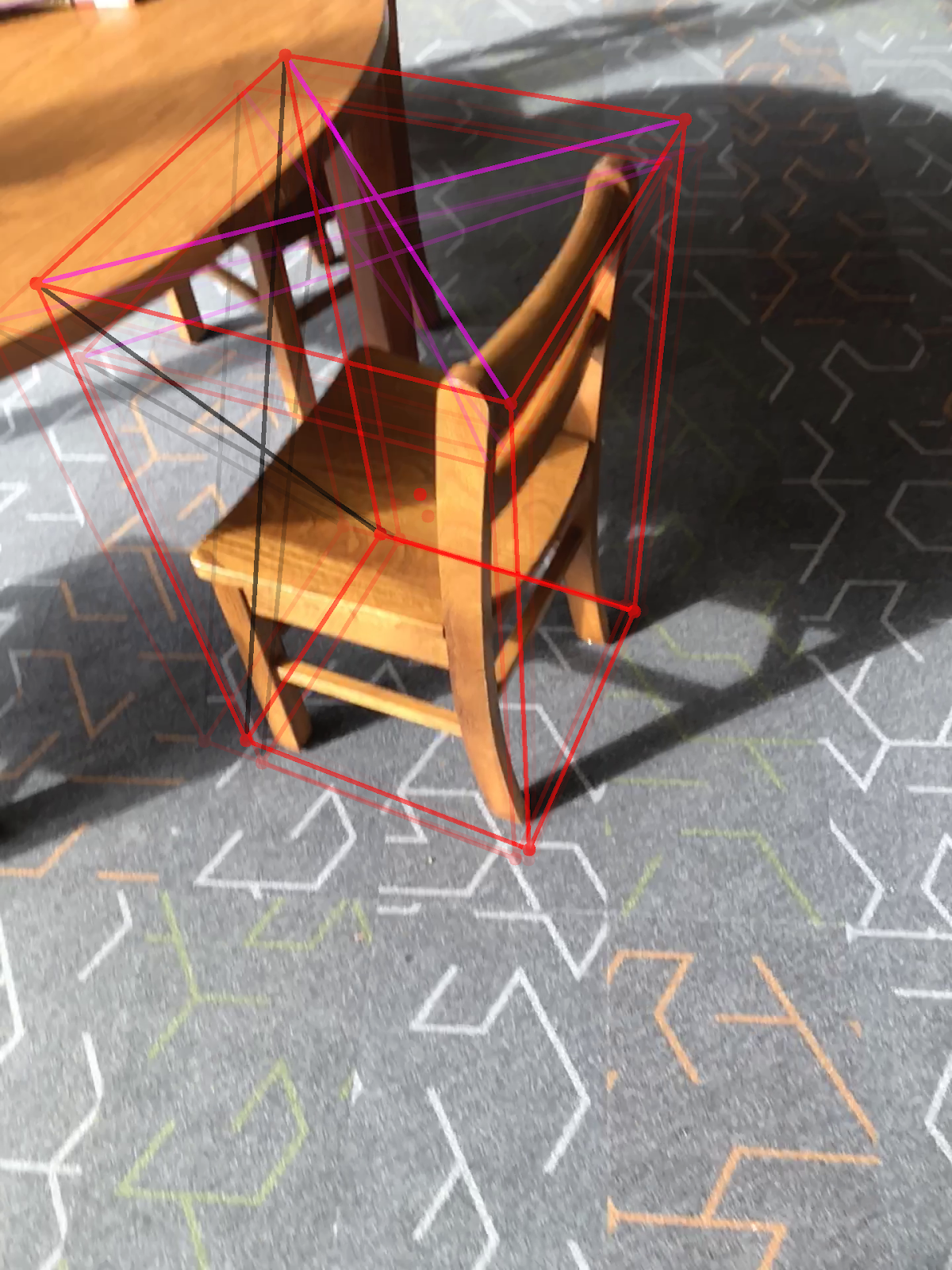}
    \includegraphics[width=0.15\textwidth]{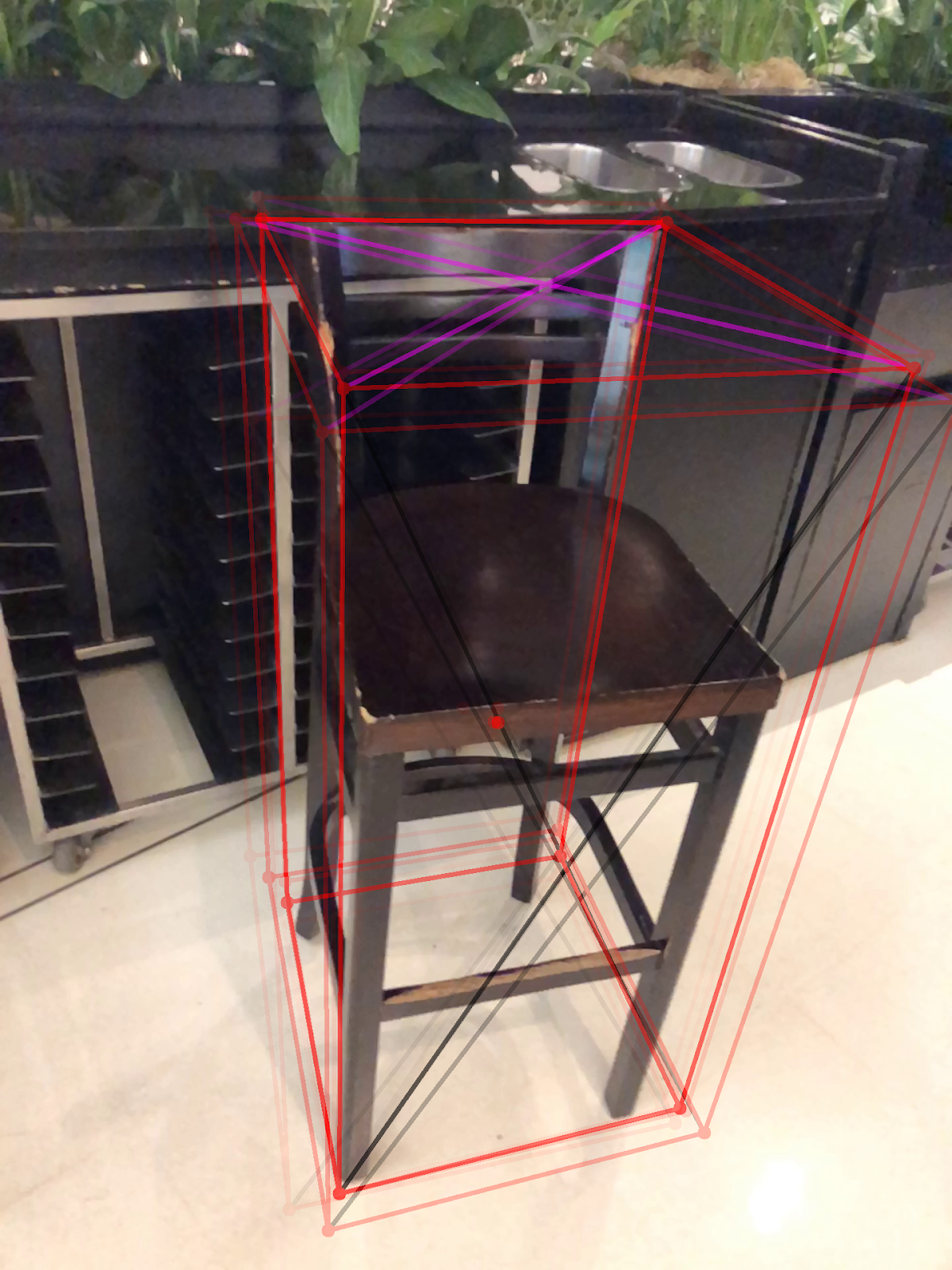}
    \includegraphics[width=0.15\textwidth]{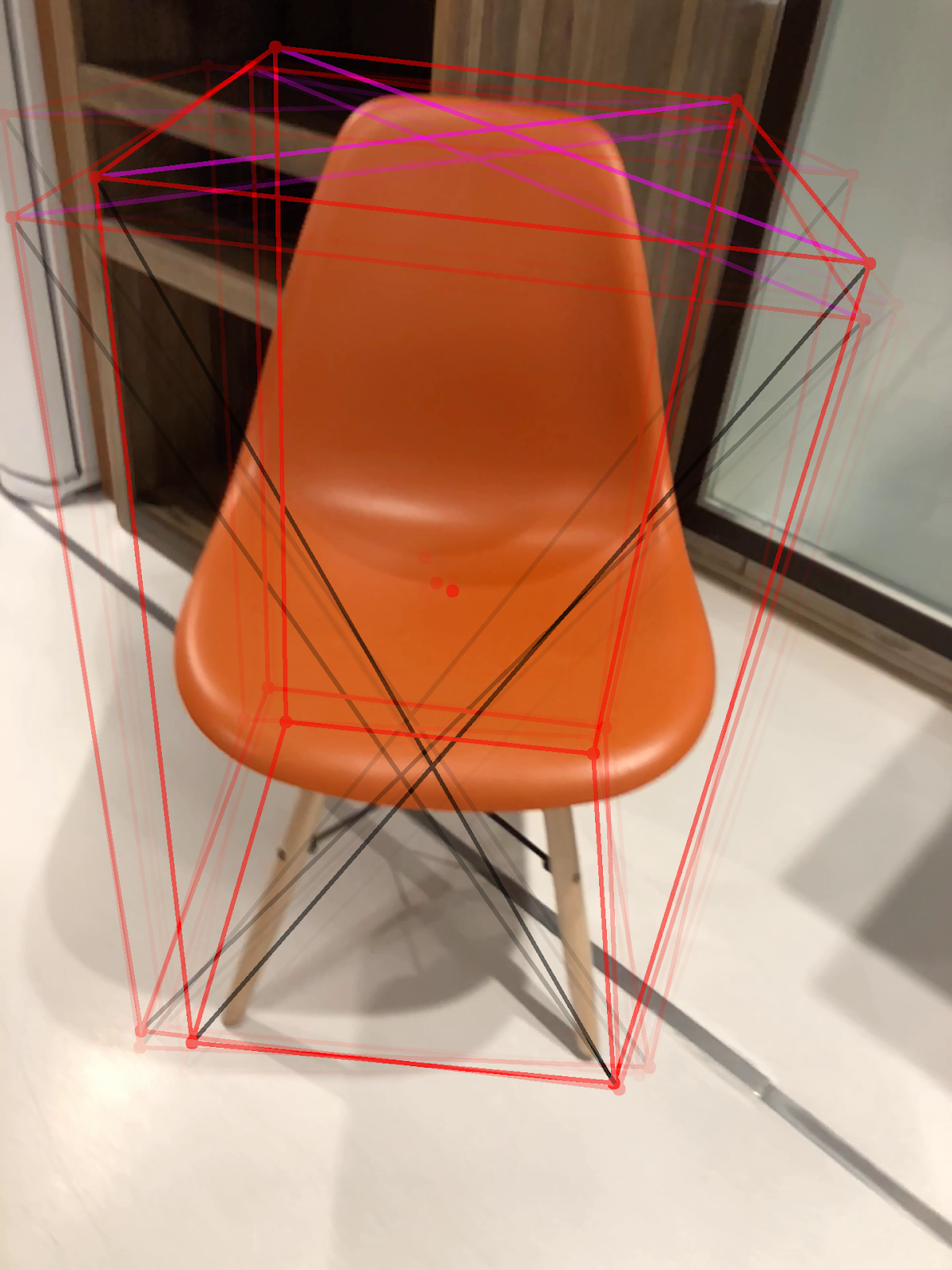}
    \includegraphics[width=0.15\textwidth]{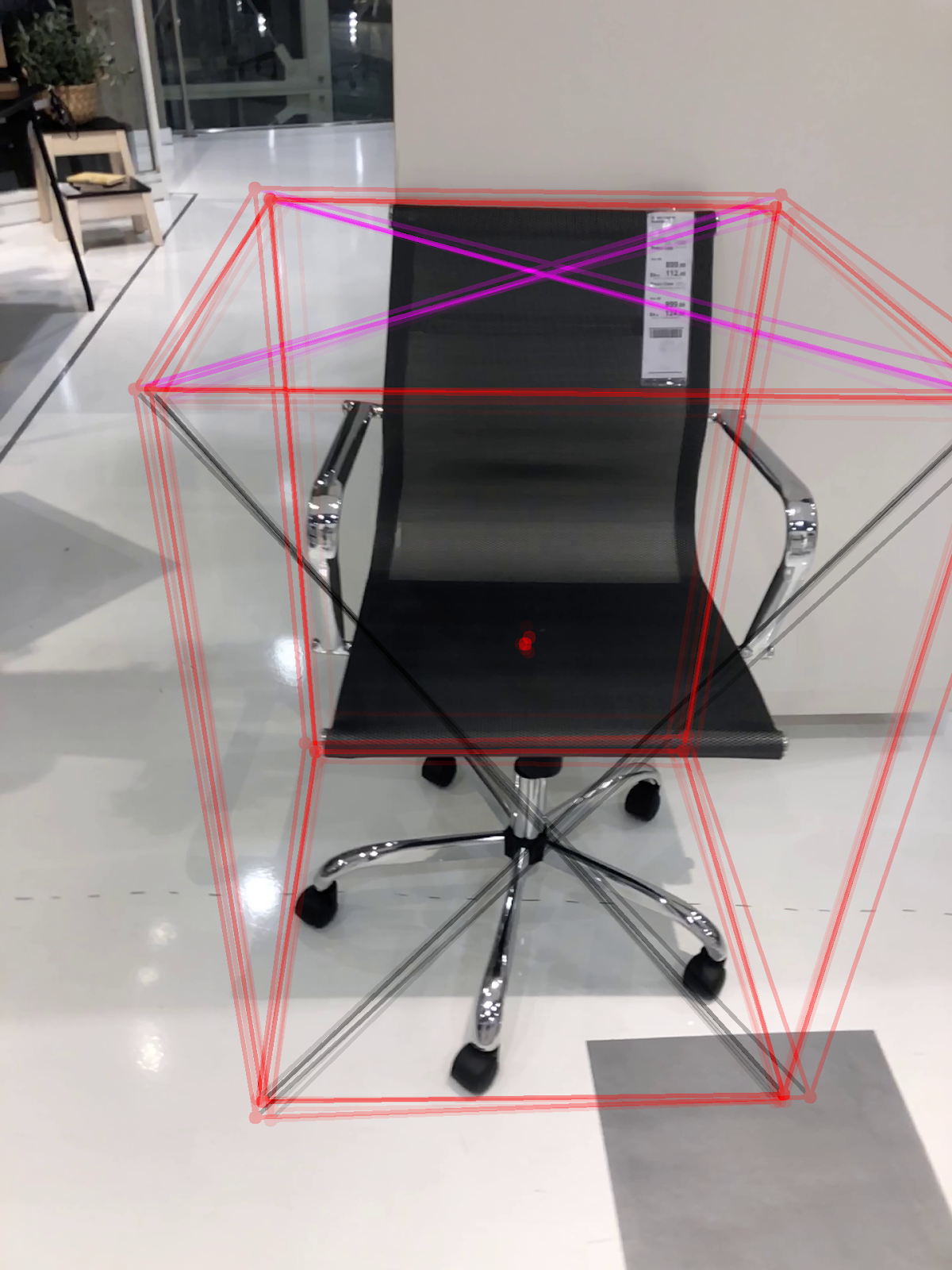}

    \includegraphics[width=0.15\textwidth]{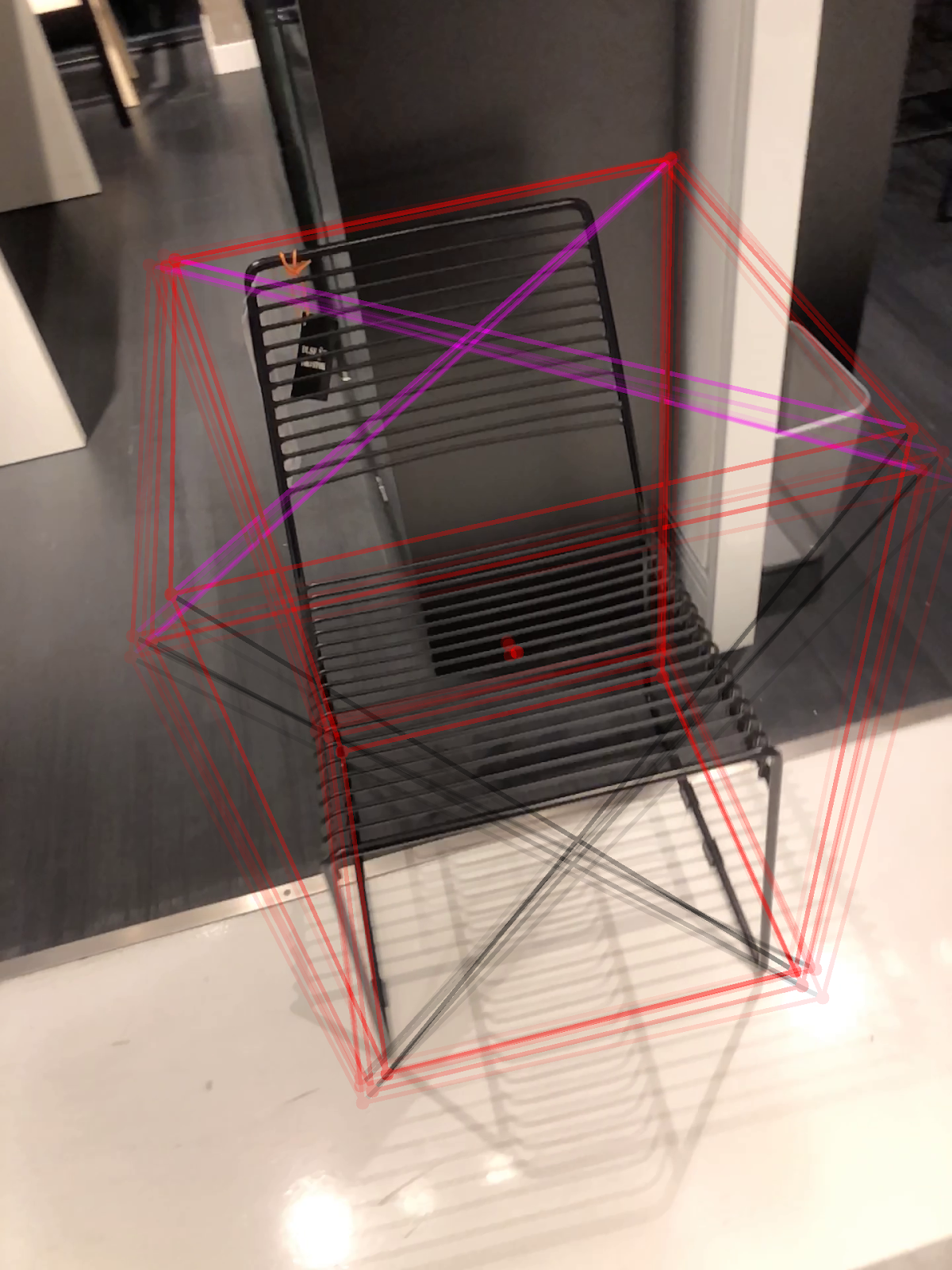}
    \includegraphics[width=0.15\textwidth]{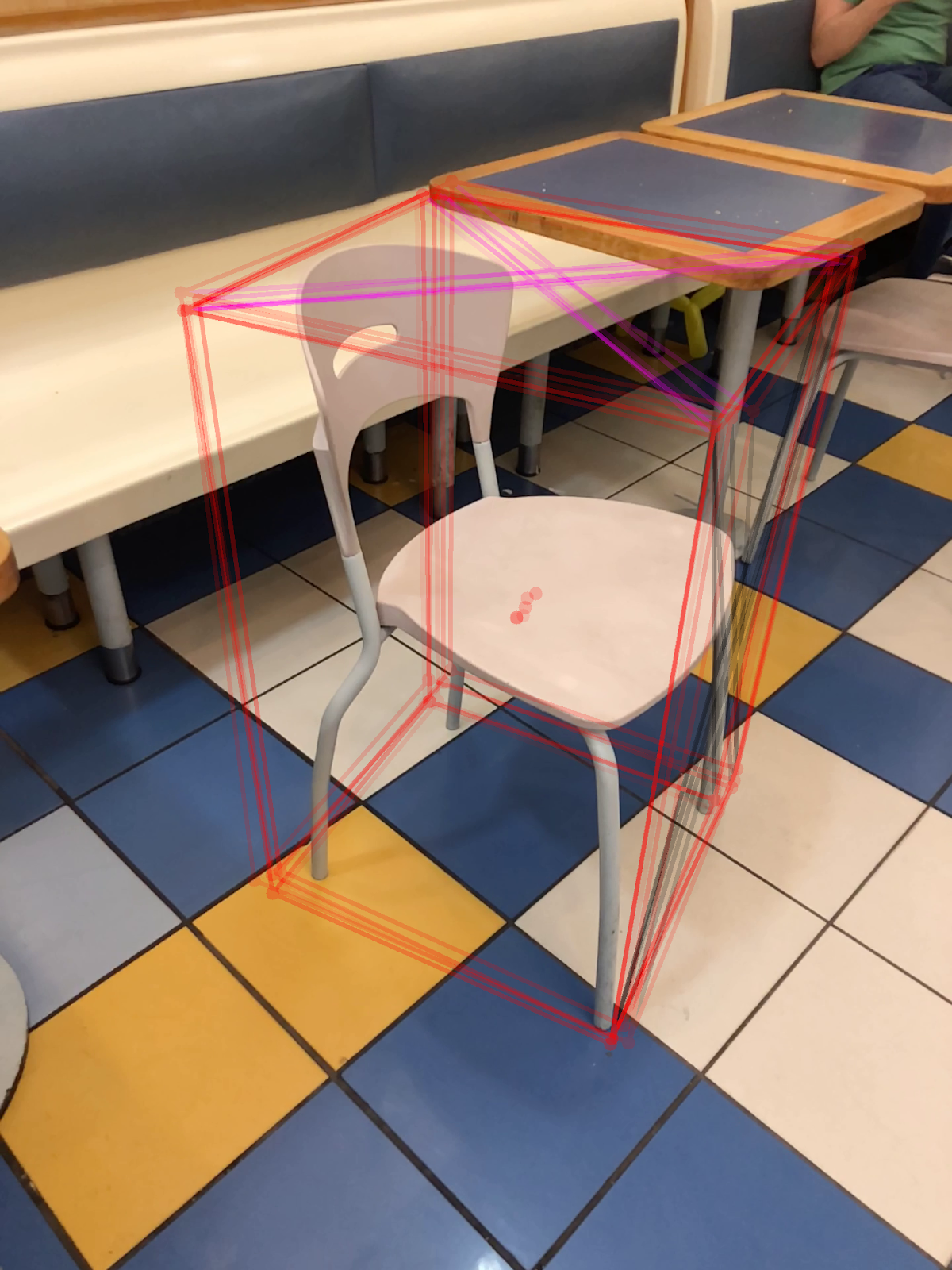}
    \includegraphics[width=0.15\textwidth]{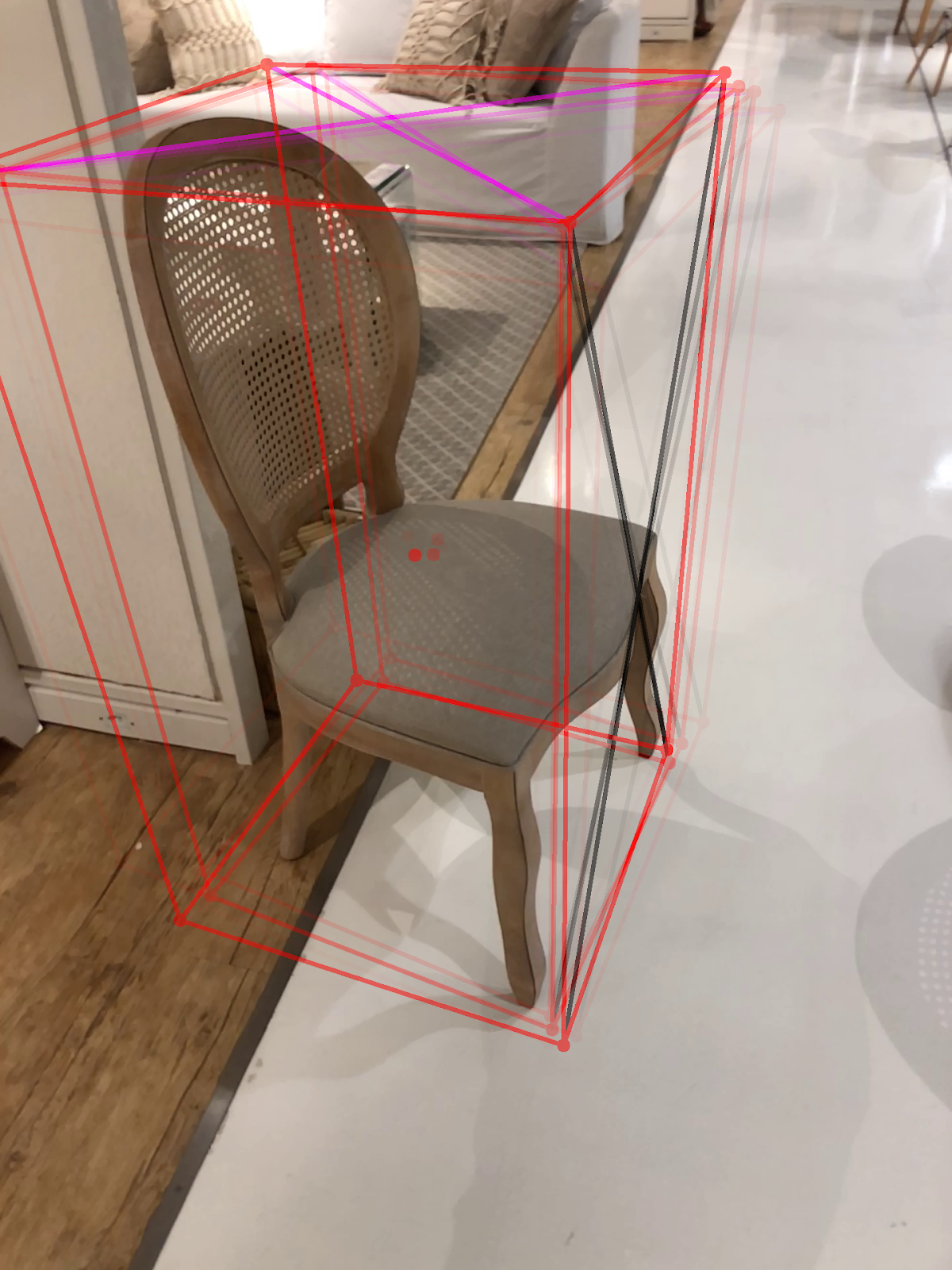}
    \includegraphics[width=0.15\textwidth]{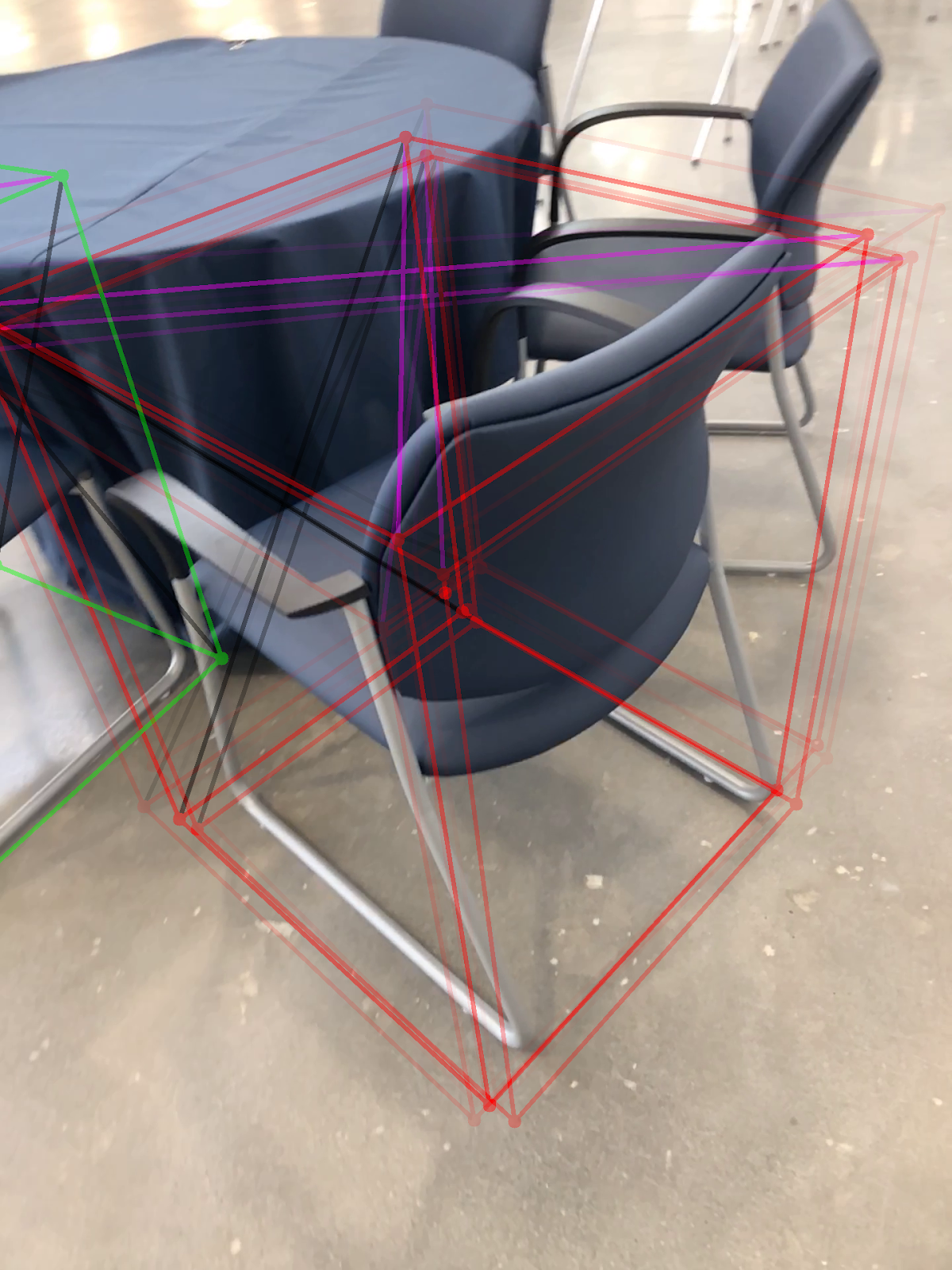}
    \includegraphics[width=0.15\textwidth]{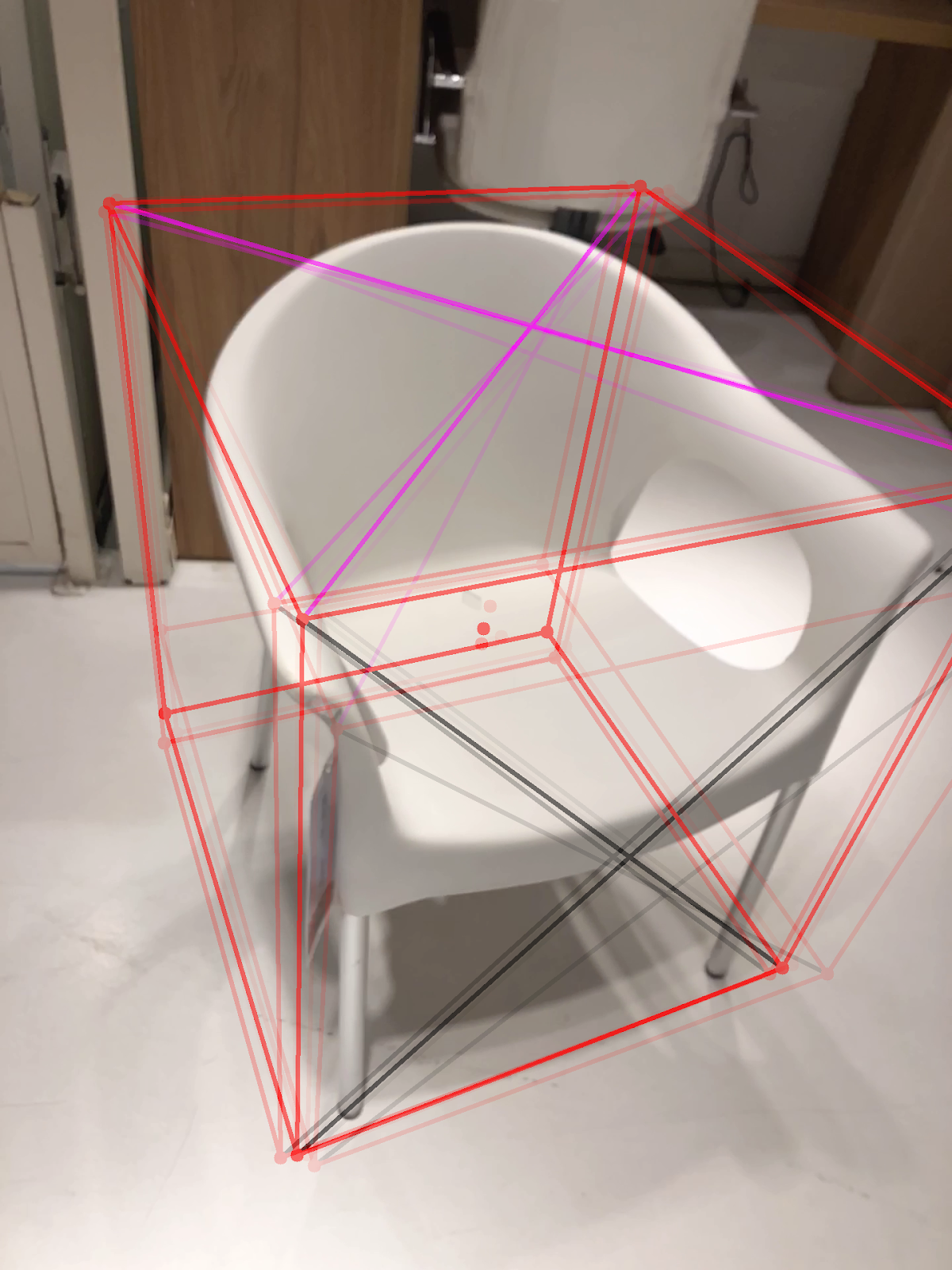}
    
\caption{The overlay of the 3D bounding boxes annotated by different annotators shows the annotations from different raters are very close.}
\label{fig:variance}
\end{figure*}

The accuracy of our annotation hinges on two factors: 1) the amount of drift in the estimated camera pose throughout the captured video, and 2) the accuracy of the raters annotating the 3D bounding box. We empirically observed our videos drift $<2\%$ in length. To further reduce the drift, we usually capture shorter sequences below $10sec$. Figure~\ref{fig:length} shows the distribution of the length of the videos in our dataset. To evaluate the accuracy of the rater, we asked eight annotators to re-annotate same sequences. A few samples are shown in Figure~\ref{fig:variance}. Overall for the chairs, the standard deviation for the chair orientation, translation, and scale was $4.6\degree$, $1cm$, and $4cm$, respectively which demonstrates insignificant variance of the annotation results between different raters.

\section{Objectron Dataset} \label{sec:dataset}
\begin{figure*}[tbh]
    \centering
    \begin{subfigure}[t]{\textwidth}
        \includegraphics[width=\textwidth]{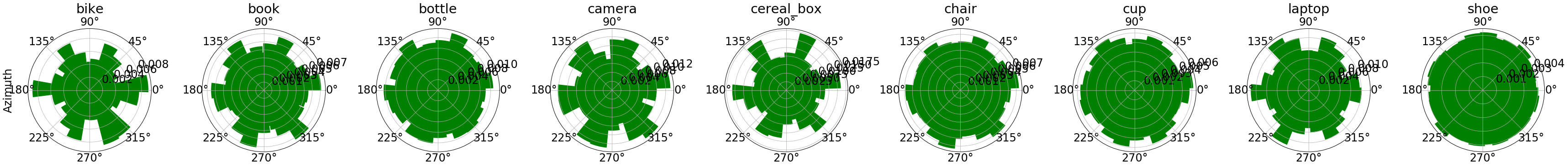}
    \end{subfigure}
    \hspace{5pt}
    \begin{subfigure}[t]{\textwidth}
        \includegraphics[width=\textwidth]{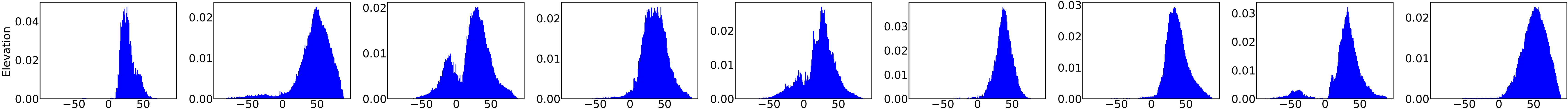}
    \end{subfigure}
    \caption{View-point distribution of samples per object category. The top row shows the azimuth distribution in polar graph, and the bottom row denotes the elevation distribution.}
    \label{fig:viewpoint}
\end{figure*}

\begin{table*}[tbh]
\centering
\begin{tabular}{|l|l|l|l|l|l|l|l|l|l|}
\hline
class                     & bike & book & bottle & camera & cereal\_box & chair & cup  & laptop & shoe \\ \hline
No. videos                & 476  & 2024 & 1928   & 815    & 1609        & 1943  & 2204 & 1473   & 2116 \\ \hline
No. frames                & 150k & 576k & 476k   & 233k   & 396k        & 488k  & 546k & 485k   & 557k \\ \hline
No. instances             & 487  & 2084 & 2014   & 931    & 1638        & 2362  & 2378 & 1710   & 3500 \\ \hline 
Avg instance per video    & 1.03 & 1.02 & 1.04   & 1.14   & 1.01        & 1.21  & 1.07 & 1.16   & 1.65 \\ \hline
\end{tabular}
\caption{Per-category statistics of Objectron dataset.}
\label{tab:stats}
\end{table*}

In this section, we describe the details of our Objectron dataset and provide some statistics. The dataset contains videos in 9 categories: bikes, books, bottles, cameras, cereal boxes, chairs, cups, laptops, and shoes. Some of these objects are non-rigid (e.g. bikes or laptops) and all of them remain stationary during the video recording.  In each video, the camera moves around the object, capturing it from different angles. In total there are $17,095$ object instances that appear in $4M$ annotated images from $14819$ annotated videos (not counting the unreleased evaluation set for future competitions). The dataset is collected from ten different countries. This is important for categories that contain texts and labels, such as books, cereal boxes and bottles. So our samples contain different languages as well as different local environments. Each category is divided to train and test sets. Table~\ref{tab:stats} shows detailed per-category statistics of our dataset.

Each sample contains the high-resolution image, along with the camera pose, point-cloud (from tracking), and planar surfaces in the environment. The data also contains manually annotated 3D bounding boxes for each object, which describe the object’s orientation, translation, and size relative to the camera pose. Our coordinate system follows a left-hand rule, where $+y$ axis is up. From the pose, we also compute the object's bounding box 3D keypoints,  projected 2D keypoints, as well as azimuth and elevation. Furthermore, each frame in the video contains the camera pose, and the camera's projection and view matrices. 

The annotation of the object contains its rotation, translation w.r.t. the camera center as well as the object's scale. To get a better understanding of the viewpoint distribution, we computed the azimuth of each object instance w.r.t the camera center. Here azimuth 0 degree indicates the object is being viewed from the front. Figure ~\ref{fig:viewpoint}-top shows the azimuth distribution for different object categories in our dataset. For some categories, there is a specific bias toward the front and top views.

\section{Baseline Experiments and Evaluations} \label{sec:eval}

Our dataset is released with an evaluation code to assess the performance of 3D object detection algorithms using various metrics. The evaluation code computes the average precision for multiple 2D and 3D metrics such as 3D IoU, 2D projection error, view-point error, polar and azimuth error, and rotation error.

Except for the 3D IoU, the other metrics are fairly standard and we refer the readers to our code as well as other references for further details. In this section, we explain our implementation of 3D IoU metric, which to the best of our knowledge, is novel. 

\subsection{3D Intersection Over Union} \label{sec:3diou}
Intersection over Union (IoU), also known as {\it Jaccard} index, is an evaluation metric for computer vision tasks such as object detection, segmentation, and tracking. It measures how close the prediction bounding boxes are to the ground truth. The IoU takes two boxes as input: the predicted box (by the model) and the ground truth box and computes the intersection volume of the two boxes. The IoU metric is invariant under change in the scale, as well as any rigid transformations that belong to the SE3 group. Therefore if we change the coordinate system, or rotate, translate, and scale both boxes by the same transformation, their IoU does not change. This property becomes useful later. The IoU is a normalized metric and the it's value ranges between 0 to 1, where 1 is considered the perfect score.

For evaluation, although 3D IoU has been used in previous work, its computation is overly simplified based on some assumptions. In~\cite{Wang:2019uv}, 3D boxes are assumed to be axis-aligned. Another approach, used by autonomous datasets (e.g. in~\cite{Mousavian:2016vy} and others), is to project the 3D bounding boxes to the ground plane and then compute the intersection of the 2D projected polygons. Then the intersection volume is estimated by multiplying the area of the 2D intersecting polygon with the height of the 3D bounding boxes. Although this approach works for vehicles on the road, it has two limitations: 1) The object should sit on the same ground plane, which limits the degrees of freedom of the box from 9 to 7. The box only has freedom in yaw, and the roll and pitch are set to 0. 2) it assumes the boxes have the same height. For the Objectron datasets, these assumptions do not hold.
 
We propose an algorithm for computing accurate 3D IoU values for general 3D-oriented boxes. First, we compute the intersection points between the faces of the two boxes using the Sutherland-Hodgman Polygon clipping algorithm~\cite{Ericson04}. Let $x$ denote the predicted box and $y$ denote the annotation label. To compute the intersecting points between the boxes $x$ and $y$, first transform both boxes using the inverse transformation of the box $x$. The transformed box $x$ will be axis-aligned and centered around the origin while the box $y$ is brought to the coordinate system of the box $x$ and remains oriented. Volume remains invariant under rigid-body transformation. We can compute the intersecting points in the new coordinate system and estimate the volume from the transformed intersection points. Using this coordinate system allows for more efficient and simpler polygon clipping against boxes since each surface is perpendicular to one of the coordinate axes. 

\begin{figure}[h]
    \centering
    \begin{subfigure}[t]{0.2\textwidth}
        \includegraphics[width=\textwidth]{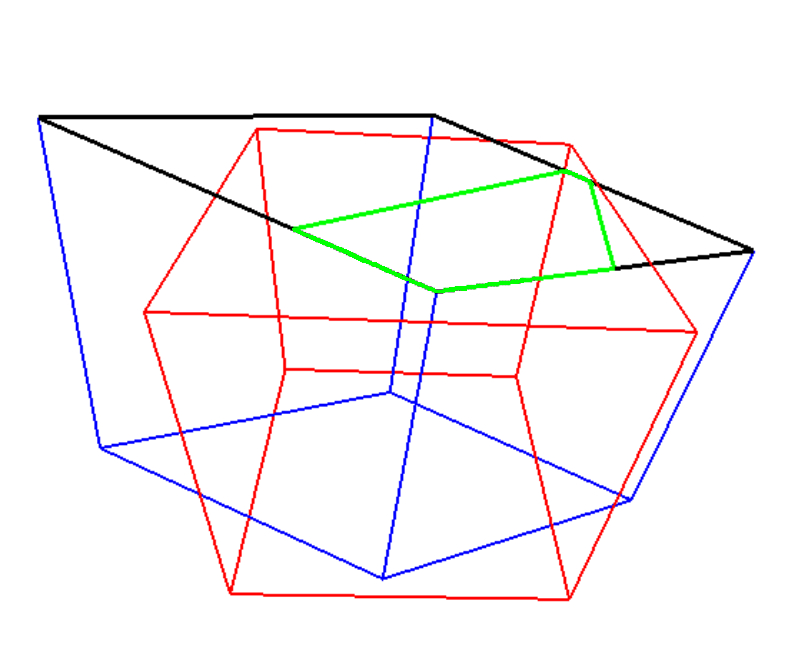}
        \caption{Clip a polygon from two boxes using the Sutherland-Hodgman algorithm.}
        \label{fig:poly_clipping}
    \end{subfigure}
    \hspace{5pt}
    \begin{subfigure}[t]{0.2\textwidth}
        \includegraphics[width=\textwidth]{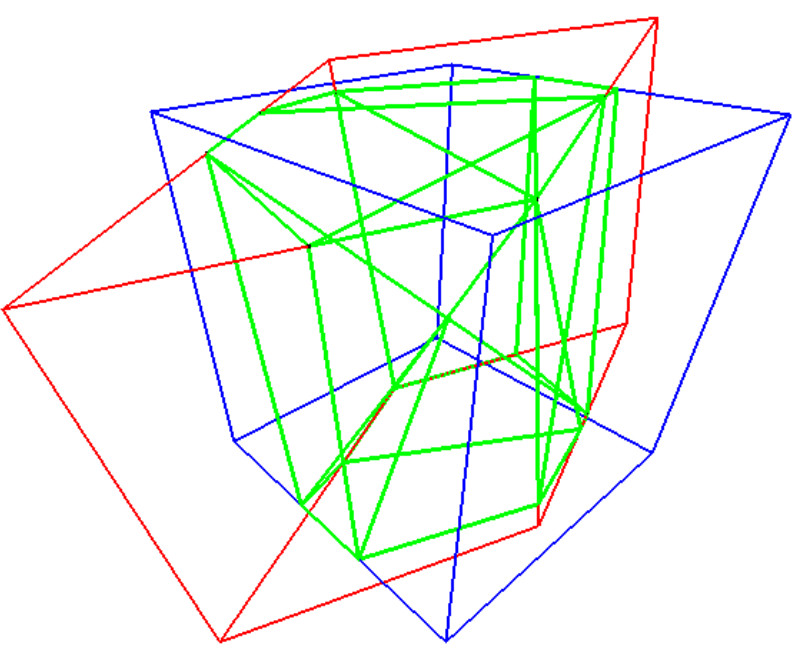}
        \caption{Compute intersection volume (green) using the convex hull algorithm.}
        \label{fig:intersection_points}
    \end{subfigure}
    \caption{Accurate computation of 3D IoU using polygon-clipping algorithm.}
    \label{fig:iou3d}
\end{figure}

Next, we clip each face in box $y$, which is a convex polygon, against the axis-aligned box $x$. There is a well-known polygon clipping algorithm in computer graphics~\cite{Ericson04}, where each world polygon is clipped against the camera frustum to determine the rendered environment. We use a robust Sutherland-Hodgman algorithm to perform the clipping.
To clip a polygon against a plane, the edges of the plane are hypothetically extended to infinity. We iterate over each edge in the polygon in clockwise order and determine whether that edge intersects with any faces in the axis-aligned box $x$. For each vertex in the box $y$, we check whether any of them are inside the box $x$. We add those vertices to the intersecting vertices as well. We repeat the whole process swapping the box $x$ and $y$. We refer the readers to ~\cite{Ericson04} for the details of the polygon clipping algorithm. Figure~\ref{fig:poly_clipping} shows an example of a polygon clipping.

The volume of the intersection is computed by the convex hull of all the clipped polygons, as shown in Figure~\ref{fig:intersection_points}. Finally, the IoU is computed from the volume of the intersection and volume of the union of two boxes. We are releasing the evaluation metrics source code along with the dataset.

\begin{figure}[h]
    \centering
    \begin{subfigure}[t]{0.2\textwidth}
        \includegraphics[width=\textwidth]{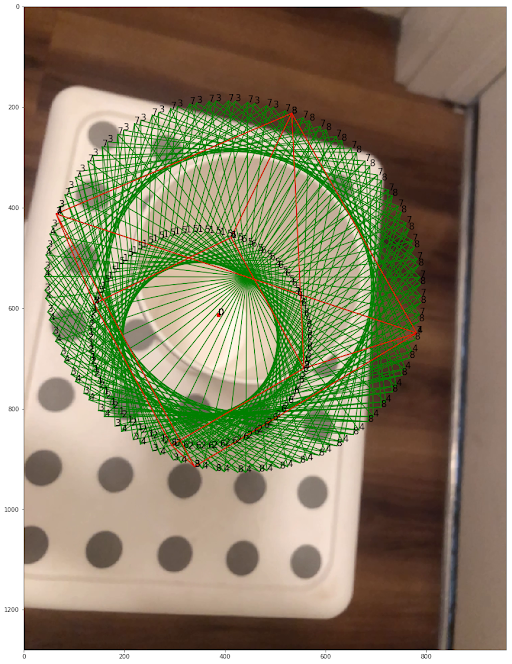}
    \end{subfigure}
    \begin{subfigure}[t]{0.2\textwidth}
        \includegraphics[width=\textwidth]{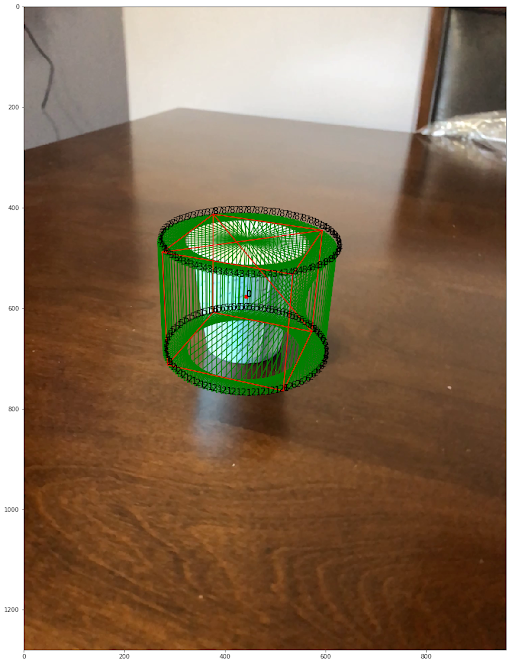}
    \end{subfigure}
    \caption{3D IoU computation for symmetric objects: Rotating the bounding box along the $Y$ axis of symmetry to maximize 3D IoU.}
    \label{fig:iou3d_sym}
\end{figure}

For symmetric objects, such as bottles or cups, the 3D IoU metric is not well-defined. In those instances, we rotate the estimated bounding box along the symmetry axis and evaluate each rotated instance, then pick the bounding box that maximizes the IoU. Figure~\ref{fig:iou3d_sym} shows an example for asymmetric cup.

\subsection{Baselines for 3D object detection} \label{sec:baseline}

\begin{figure*}[tb!]
    \centering
    \begin{subfigure}[t]{0.3\textwidth}
        \includegraphics[width=\textwidth]{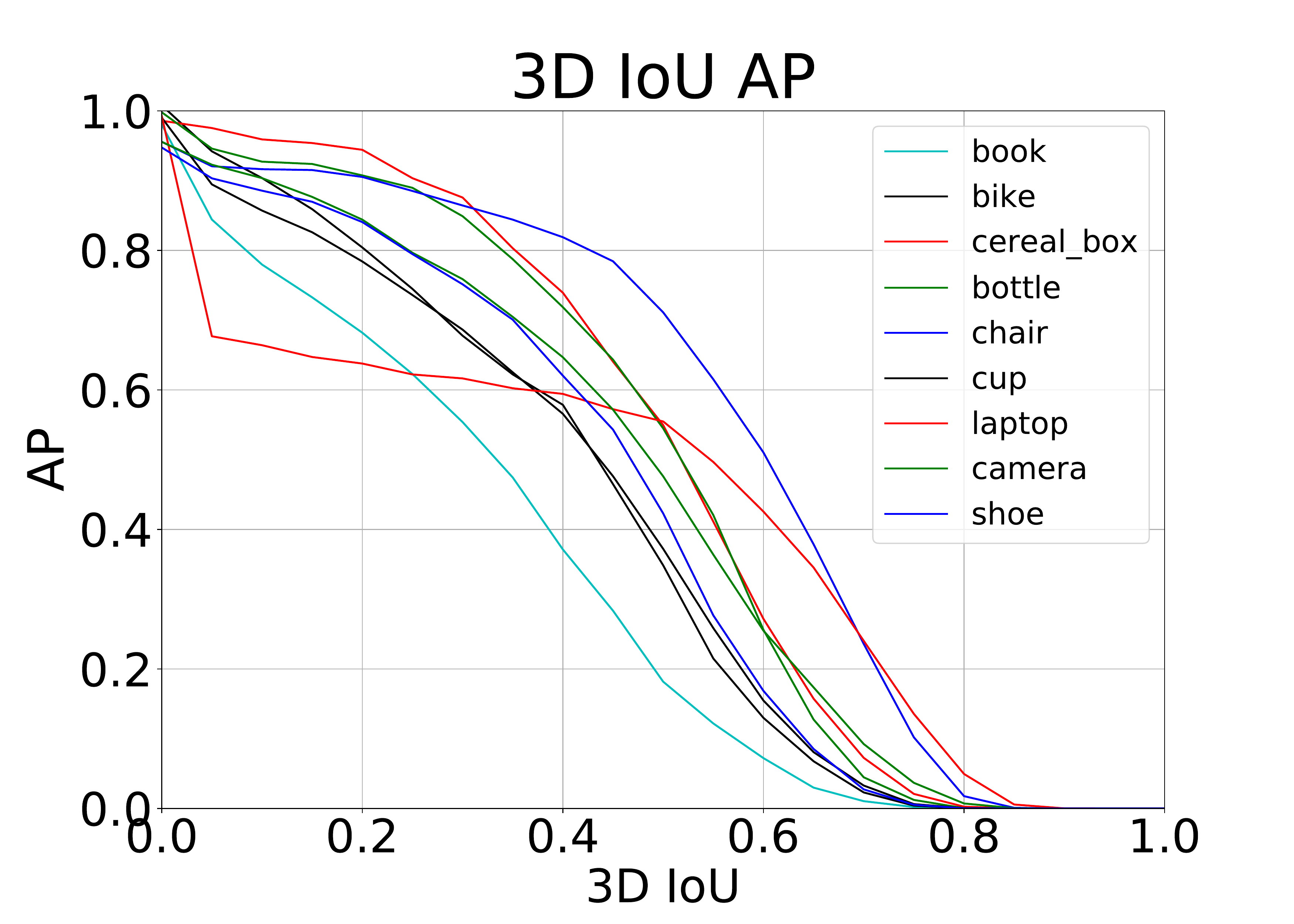}
    \end{subfigure}
    \begin{subfigure}[t]{0.3\textwidth}
        \includegraphics[width=\textwidth]{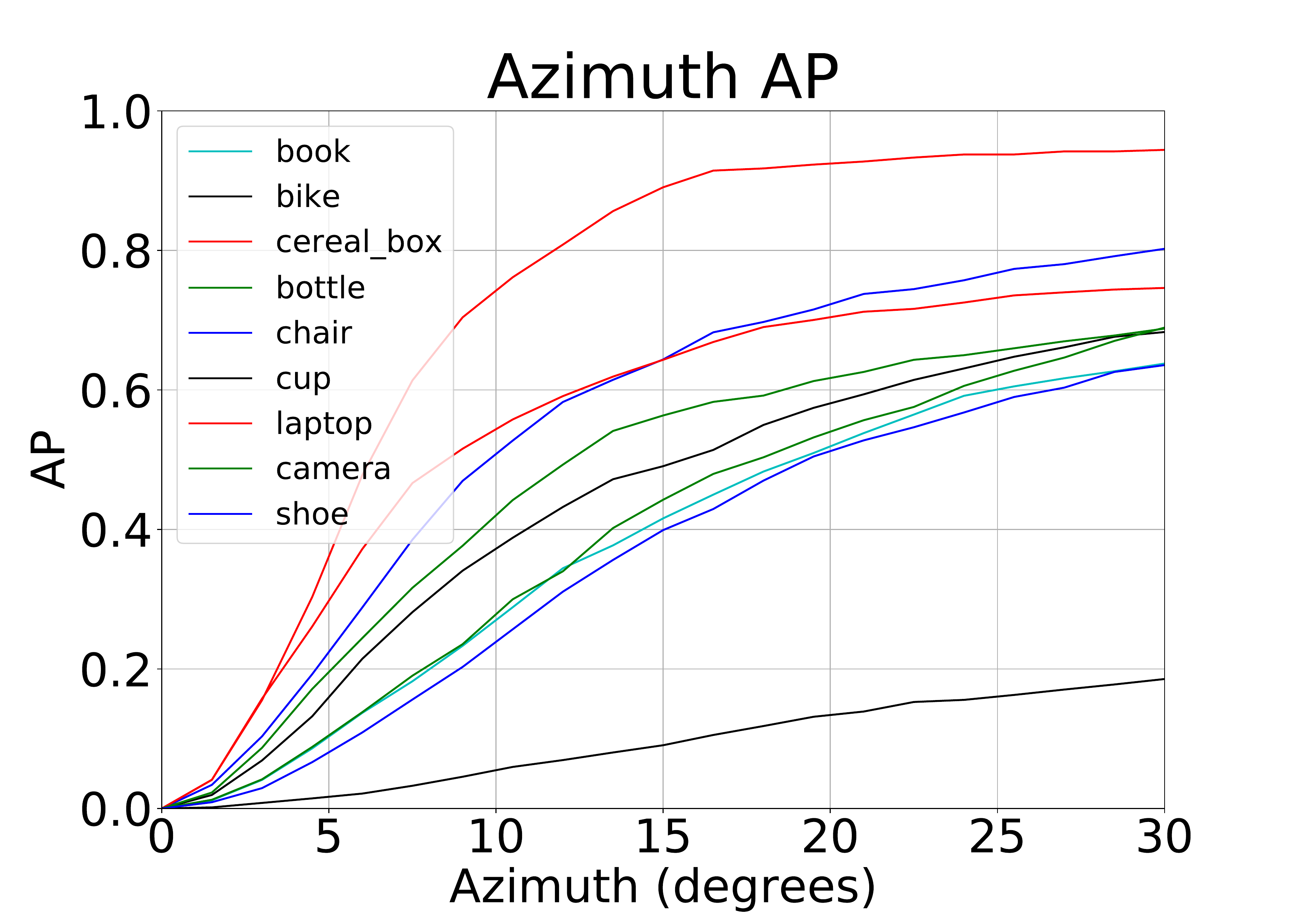}
    \end{subfigure}
    \begin{subfigure}[t]{0.3\textwidth}
        \includegraphics[width=\textwidth]{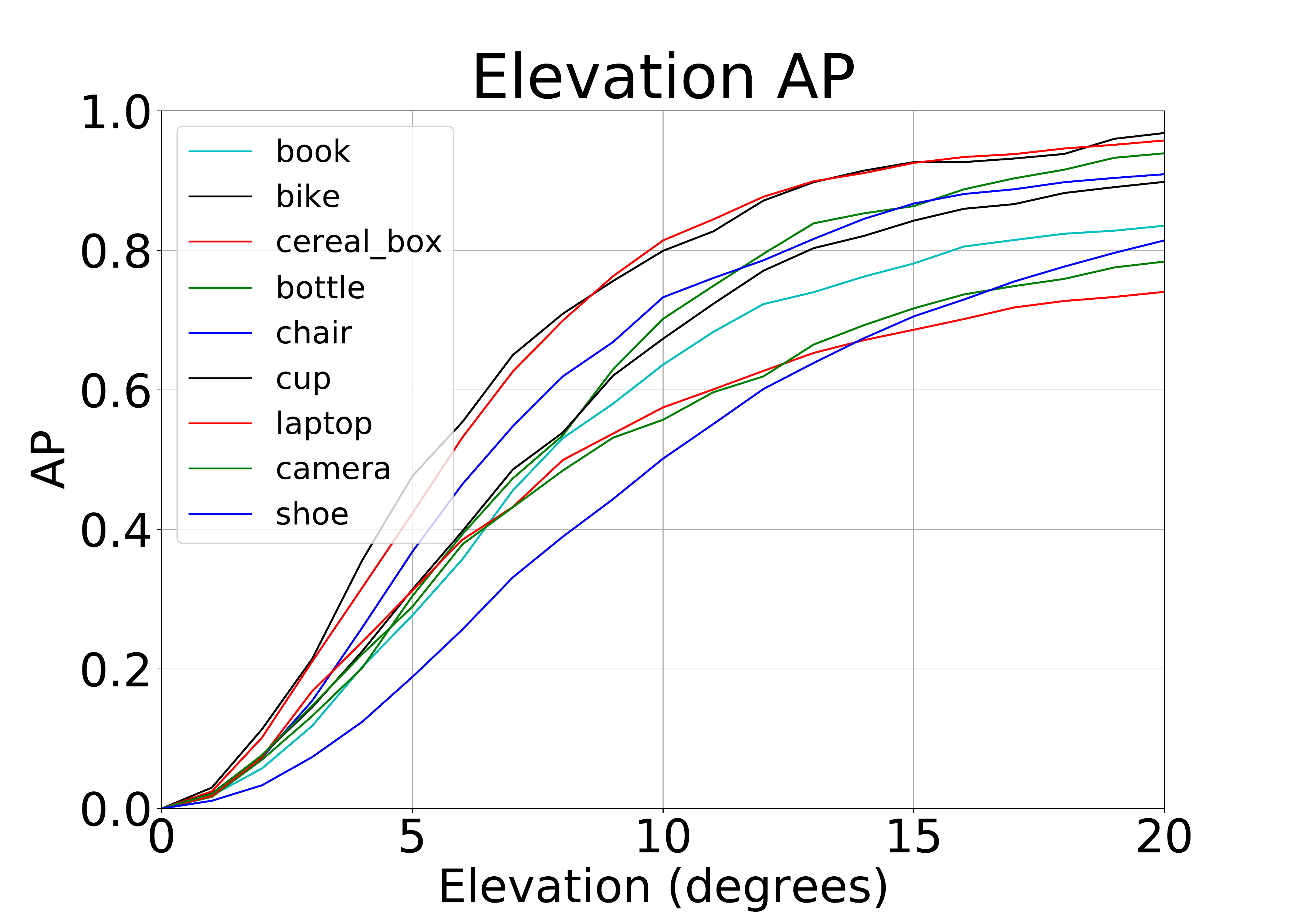}
    \end{subfigure}
    \caption{Evaluation of MobilePose network\cite{Hou:2020} on the Objectron dataset}
    \label{fig:mobilepose_eval}
\end{figure*}

\begin{figure*}[tb!]
    \centering
    \begin{subfigure}[t]{0.3\textwidth}
        \includegraphics[width=\textwidth]{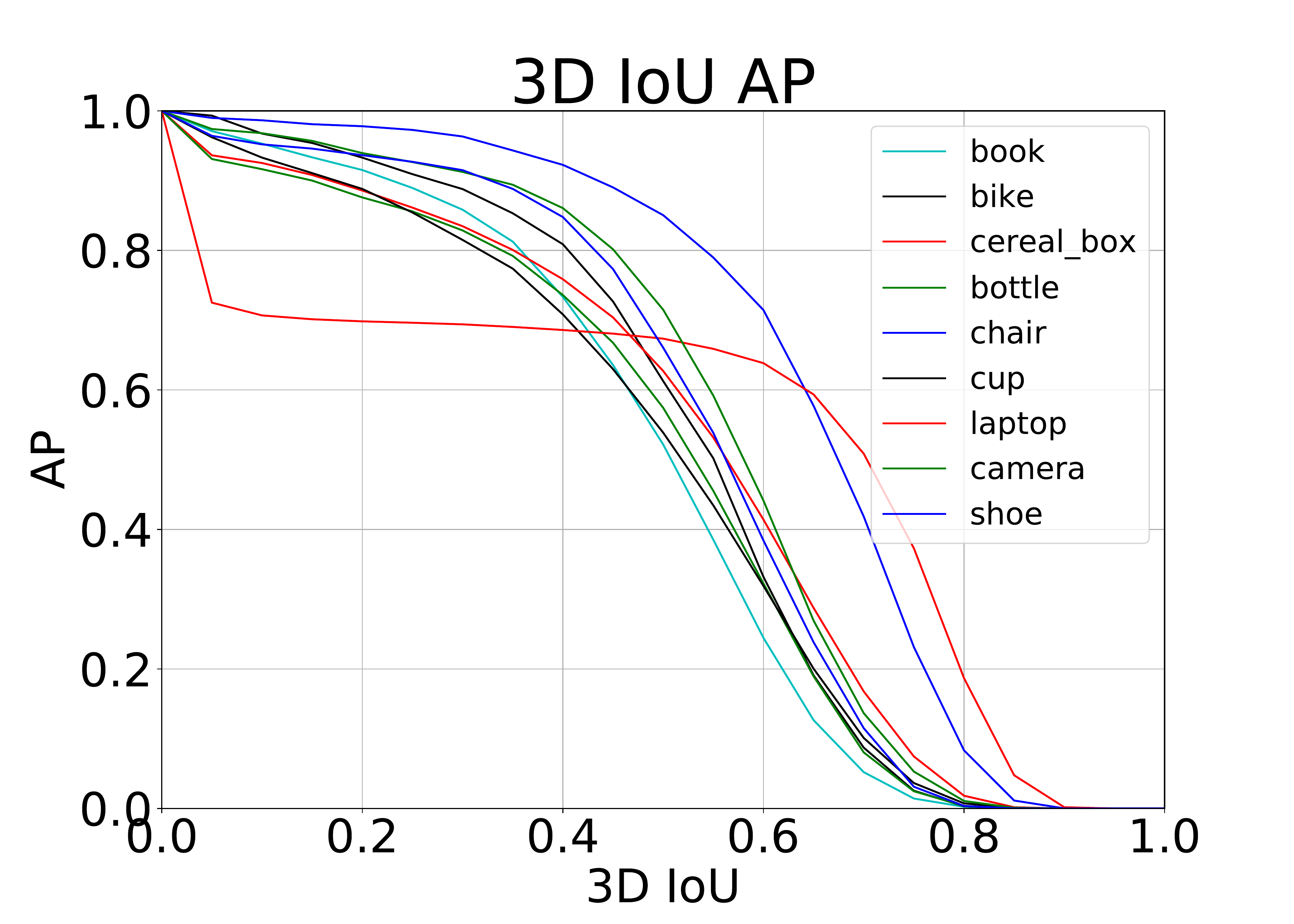}
    \end{subfigure}
    \begin{subfigure}[t]{0.3\textwidth}
        \includegraphics[width=\textwidth]{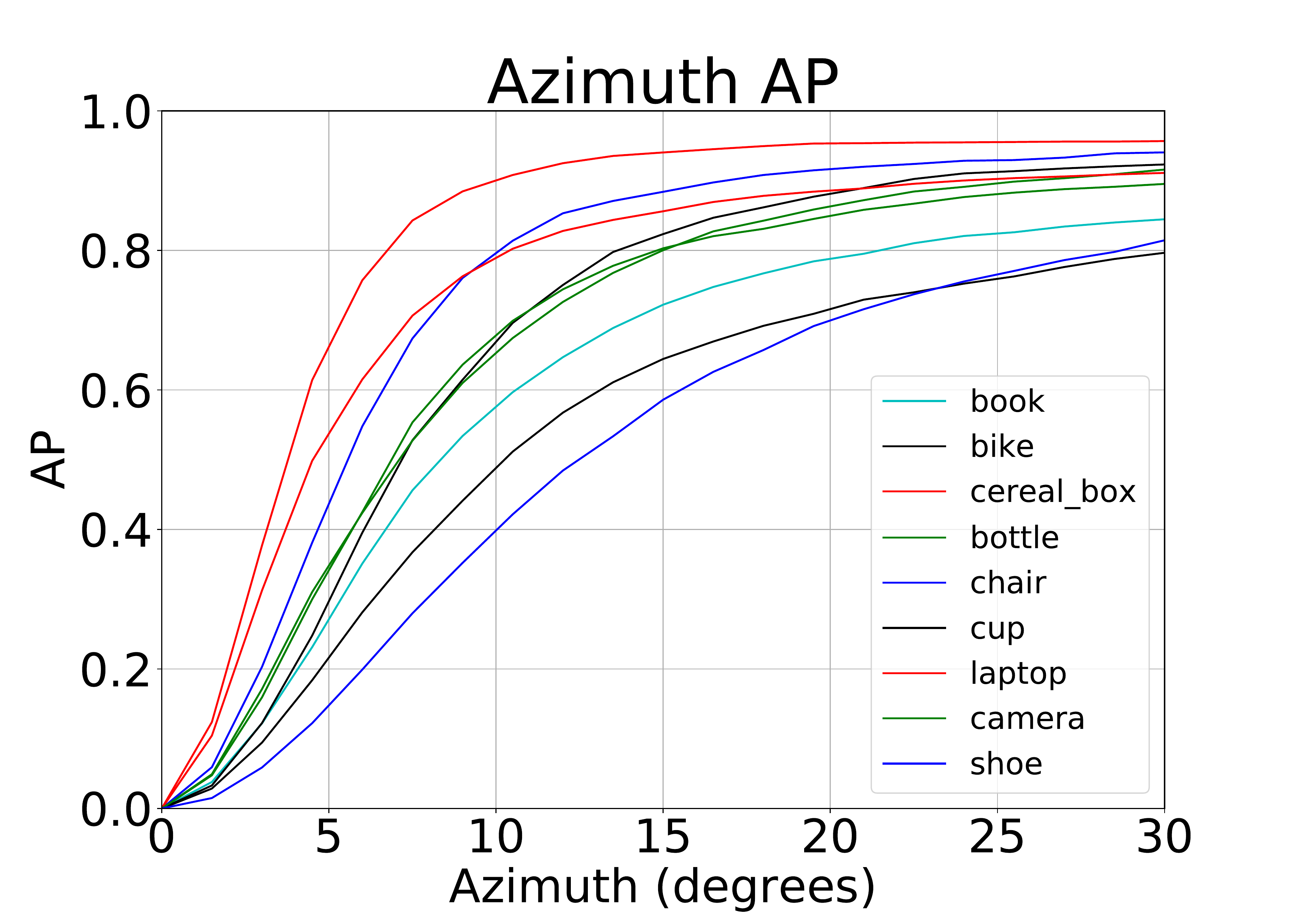}
    \end{subfigure}
    \begin{subfigure}[t]{0.3\textwidth}
        \includegraphics[width=\textwidth]{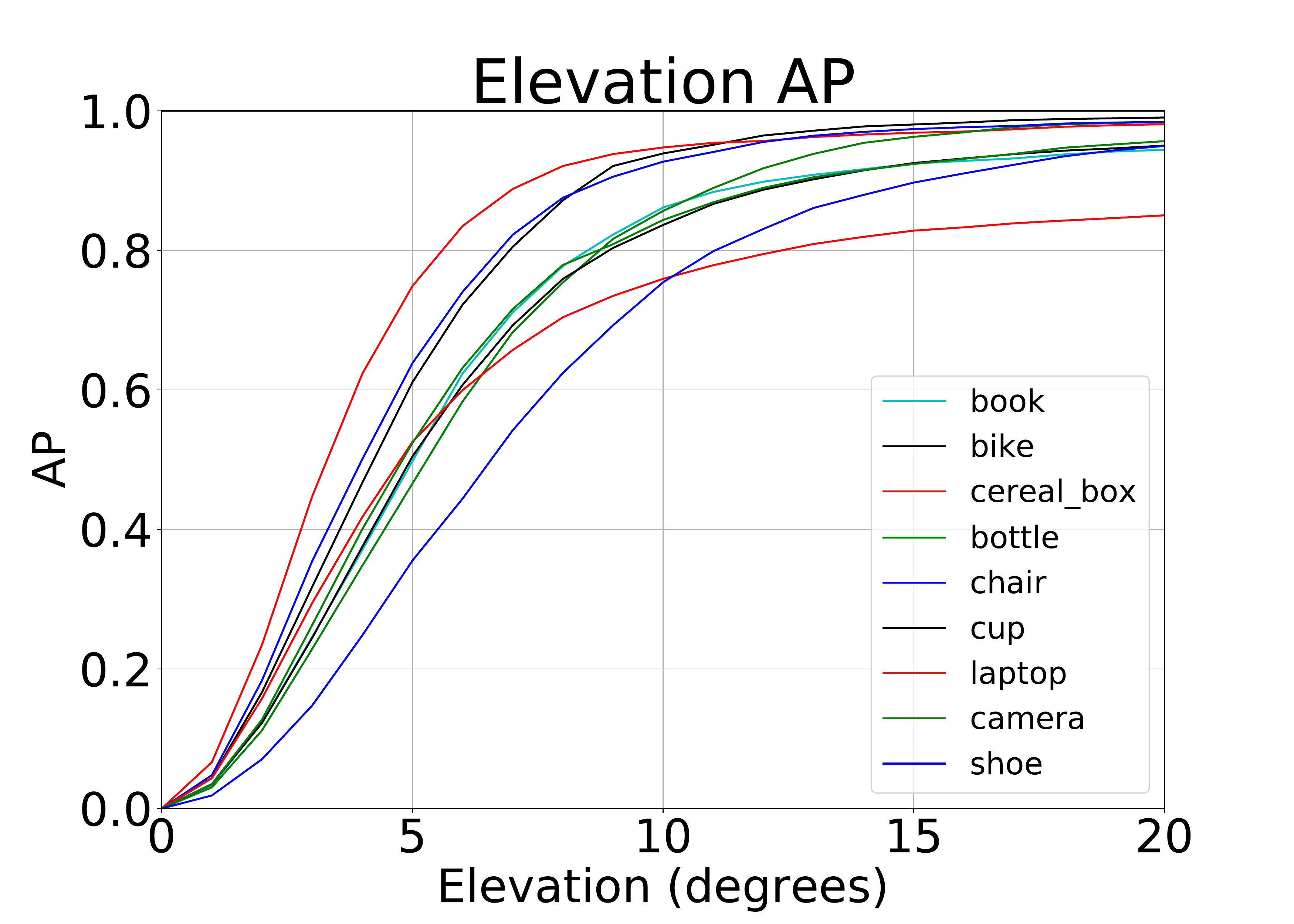}
    \end{subfigure}
    \caption{Evaluation of two-stage network on the Objectron dataset}
    \label{fig:mesh2_eval}
\end{figure*}

\begin{table*}[tb!]
    \begin{subtable}[h]{\textwidth}
        \centering
            \begin{tabular}{|l|l|l|l|l|l|l|l|l|l|}
            \hline
                Model                      & bike   & book   & bottle & camera & cereal\_box & chair  & cup    & laptop & shoe   \\ \hline
                MobilePose \cite{Hou:2020} & 0.3486 & 0.1818 & 0.5449 & 0.4762 & 0.5496      & 0.7112 & 0.3722 & 0.5548 & 0.4230 \\ \hline
                Two-stage                  & 0.6127 & 0.5218 & 0.5744 & 0.8016 & 0.6272      & 0.8505 & 0.5388 & 0.6735 & 0.6606 \\ \hline
            \end{tabular}
       \caption{Average precision at 0.5 3D IoU metric for different categories.}
       \label{tab:eval_iou}
    \end{subtable}
    \newline
    \vspace*{0.15in}
    \newline
    \begin{subtable}[h]{\textwidth}
        \centering
            \begin{tabular}{|l|l|l|l|l|l|l|l|l|l|}
                \hline
                Model                      & bike   & book   & bottle & camera & cereal\_box & chair  & cup    & laptop & shoe   \\ \hline
                MobilePose \cite{Hou:2020} & 0.1581 & 0.0840 & 0.0818 & 0.0773 & 0.0454      & 0.0892 & 0.2263 & 0.0736 & 0.0655 \\ \hline
                Two-stage                  & 0.0828 & 0.0477 & 0.0405 & 0.0449 & 0.0337      & 0.0488 & 0.0541 & 0.0291 & 0.0391 \\ \hline
            \end{tabular}
        \caption{Mean pixel error of 2D projection of box vertices for different categories.}
        \label{tab:eval_pixel}
     \end{subtable}
     \newline
     \vspace*{0.15in}
     \newline
     \begin{subtable}[h]{\textwidth}
        \centering
            \begin{tabular}{|l|l|l|l|l|l|l|l|l|l|}
                \hline
                Model                      & bike   & book   & bottle & camera & cereal\_box & chair  & cup          & laptop & shoe   \\ \hline
                MobilePose \cite{Hou:2020} & 0.4907 & 0.4159 & 0.4426 & 0.5634 & 0.8905      & 0.6437 & {\bf 0.0907} & 0.6432 & 0.3991 \\ \hline
                Two-stage                  & 0.8234 & 0.7222 & 0.8003 & 0.8030 & 0.9404      & 0.8840 & 0.6444       & 0.8561 & 0.5860 \\ \hline
            \end{tabular}
        \caption{Average precision at $15\degree$ Azimuth error for different categories.}
        \label{tab:eval_azimuth}
     \end{subtable}
     \newline
     \vspace*{0.15in}
     \newline
     \begin{subtable}[h]{\textwidth}
        \centering
            \begin{tabular}{|l|l|l|l|l|l|l|l|l|l|}
                \hline
                Model                      & bike   & book     & bottle & camera & cereal\_box & chair  & cup    & laptop & shoe   \\ \hline
                MobilePose \cite{Hou:2020} & 0.7995 & 0.6363   & 0.7020 & 0.5573 & 0.8144      & 0.7329 & 0.6735 & 0.5750 & 0.5017 \\ \hline
                Two-stage                  & 0.9390 & 0.8616   & 0.8567 & 0.8437 & 0.9476      & 0.9272 & 0.8365 & 0.7593 & 0.7544 \\ \hline
            \end{tabular}
        \caption{Average precision at $10\degree$ Elevation error for different categories.}
        \label{tab:eval_elevation}
     \end{subtable}
     \caption{Evaluation of different baseline models for the Objectron dataset.}
     \label{tab:evaluation_results}
\end{table*}

We provide baseline results for 3D object detection and viewpoint estimation. We trained a state-of-the-art model \cite{Hou:2020} over our dataset for detecting 3D bounding boxes. MobilePose is a lightweight network that is designed to perform in real-time on mobile devices. 
We evaluate the network's output using our evaluation code and report various metrics, such as average precision for 3D IoU, 2D pixel projection error, azimuth, and elevation\footnote{The trained models and their evaluation report can be downloaded from \url{objectron.dev}}. For each category, we trained the network separately without any pre-training or hyperparameter optimization. Table~\ref{tab:evaluation_results} shows the evaluation results. We empirically observed pre-training and hyperparameter optimization can significantly improve the baseline results. Each model was trained for 100 epochs ($\sim$12 hours) on eight V100 GPUs. 

The original MobilePose network also uses the shape information obtained from synthetic data. However, we showed it also works by training purely on real data without any shape information. In our implementation, we used MobileNetV2 as backend, and added two heads to the network: 1) An attention head that creates an attention mask at the center keypoint of the 3D bounding box, and 2) a regression head, that predicts the x-y adjustment of the eight other keypoint from the center keypoint. The network predicts the nine 2D projected keypoints, which are later lifted to 3D using EPNP algorithm \cite{Hou:2020}.

We also designed a new two-stage architecture for 3D object detection. The first stage estimates a 2D crop of the object of the size $224\times224$ using SSD model\cite{Liu:2015bw}\cite{Huang:2017wn}, followed by a second stage model using EfficientNet-Lite \cite{Tan:2019ue} architecture which uses the 2D crop to regress the keypoints of the 3D bounding box. We use a similar EPnP algorithm as in~\cite{Hou:2020} to lift the 2D predicted keypoints to 3D. This network is very lightweight (5.2MB size) and runs at 83fps on Samsung S20 mobile GPU. Figure~\ref{fig:mesh2_eval} shows the evaluation results, including the average precision plot for 3D IoU, azimuth, and elevation for the two-stage model.

\begin{figure*}[tb]
    \centering
    \begin{subfigure}[t]{\textwidth}
        \includegraphics[width=\textwidth]{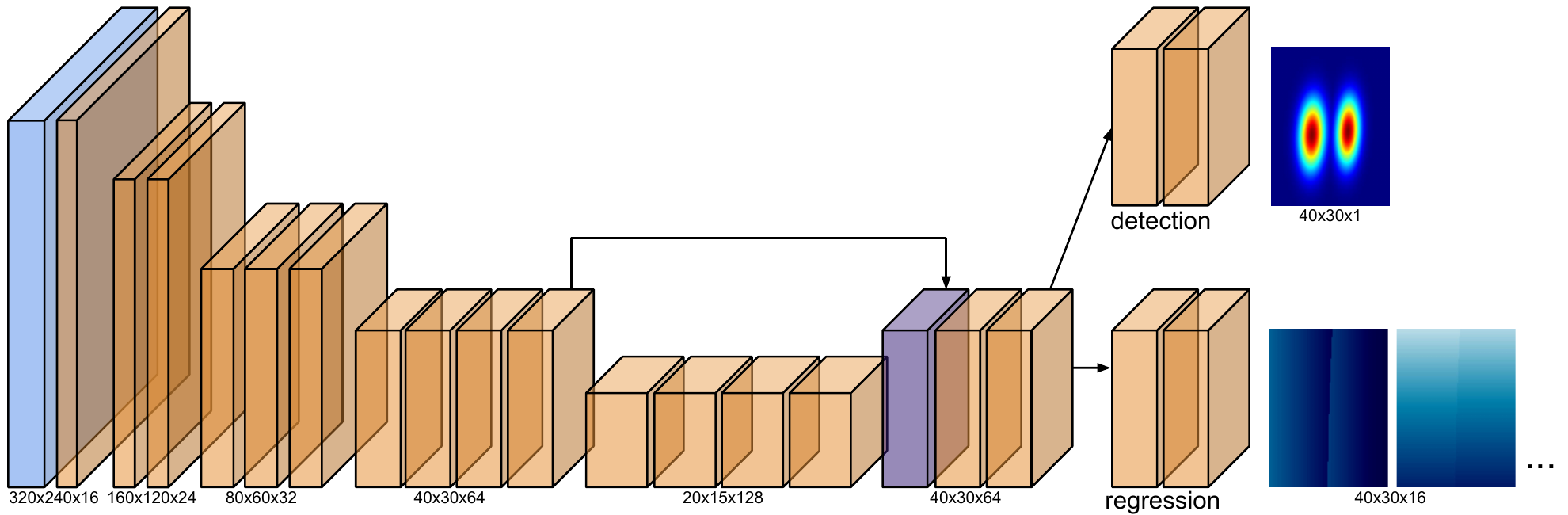}
        \caption{The architecture of the MobilePose\cite{Hou:2020} model.}
    \end{subfigure}
    \hspace{5pt}
    \begin{subfigure}[t]{\textwidth}
        \includegraphics[width=\textwidth]{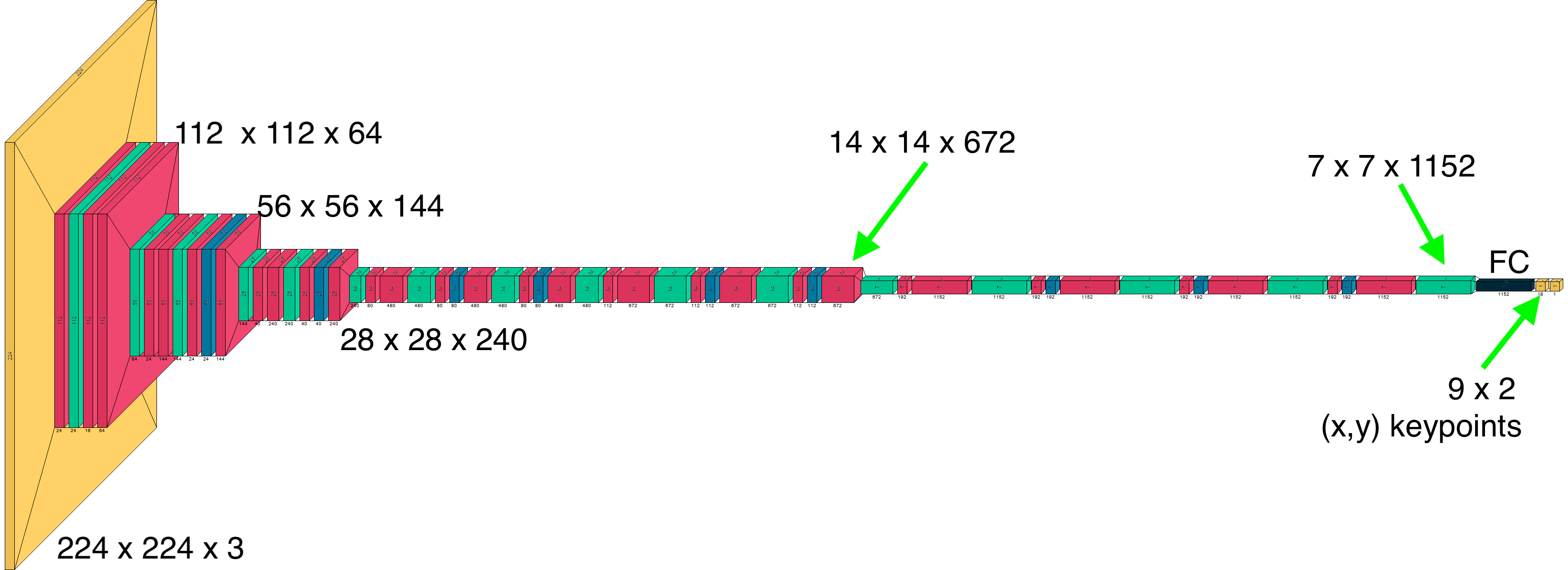}
        \caption{Architecture of the two-stage model. The red blocks are $1 \times 1$ convolutional layers, green blocks are depthwise convolutional layers, and blue blocks are addition layers for skip connection. The black block at the end is  a fully connected layer.}
    \end{subfigure}
    \caption{The baseline models used for 3D Object Detection task.}
    \label{fig:models}
\end{figure*}

The two-stage network first uses a 2D object detector (SSD network in our implementation) to detect a $224 \times 224$ crop of the object. Then the network (as shown in Figure~\ref{fig:models}) uses an EfficientNet-lite network to encode the input image to an $7 \times 7 \times 1152$ embedding vector, followed by a fully connected layer to regress the 9 2D keypoints. The network uses a similar EPnP algorithm as in~\cite{Hou:2020} to lift the 2D predicted keypoints to 3D. The hyper-parameters of the training jobs are provided in Table~3.

For the average precision, first, the detector has to detect the 3D bounding box using the center of the box, then proceed to compute the other metrics. The experimental result shows the model is more accurate in estimating elevation than azimuth because the distribution of the elevation in our dataset (Figure~\ref{fig:viewpoint}) is biased toward $45\degree$, but azimuth is uniformly distributed. In other words, in our videos, the data collector is looking down and walking around the object to capture the video. Data augmentation techniques, such as affine transformation or cropping, can change the distribution of viewpoints in the dataset and help generalization. We should also highlight how badly the network performs on estimating the rotation of the 'cup', as evident by the average precision of azimuth for the cups (Figure~\ref{fig:mobilepose_eval} and Table~\ref{tab:eval_azimuth}).

Figure~\ref{fig:models} shows the overview architecture of the models that we used as baselines. Both models are capable of achieving real-time performance on mobile devices.

\begin{table*}[tb]
\begin{tabular}{|l|l|l|}
\hline
              & MobilePose                         & Two-stage \\ \hline
Epoch         & 100                                & 250         \\
Learning rate & 1e-2, decayed to 1e-3 in 30 epoch  & 1e-2 to 1e-6 exponential annealing         \\
Batch-size    & 128                                & 64       \\
Optimizer     & adam                               &  adam         \\
Input-size    & $480 \times 640 \times 3$          &  $224 \times 224 \times 3$         \\
Input-size    & $120 \times 160$                   &  $9\times 2$          \\
Loss          & MSE on attention + $L_1$ on regression & Per vertex MSE normalized on diagonal edge length \\
Backend       & MobileNetV2                        & EfficientNet-Lite \\  \hline
\end{tabular}
\label{tab:hyp-param}
\caption{Hyper parameters for the baseline models.}
\end{table*}

\section{Details of the Objectron data format}\label{appendix:format}

The data is stored in Google storage \texttt{objectron} bucket for public access. The supplementary code for parsing and evaluation is available at \url{https://github.com/google-research-datasets/Objectron}. For each sample, the dataset provides the raw video file (in MOV file format, at 30fps, and $1440 \times 1920$ resolution, the AR Metadata, and the annotation result. The AR metadata contains the camera transformation, view, projection, and intrinsic matrix. The camera transformation contains the camera transformation from the first frame in the sequence. Furthermore, the sparse point-cloud in the world-coordinate and surface planes (including the normal and extend, boundary points, and plane alignment w.r.t. gravity vector) are provided.

For each object instance, the annotation data includes the bounding box's orientation, translation, and scale, as well as the  3D vertices in the world and camera coordinate system and their 2D projection (with depth) in the image plane. Each instance has a label string that corresponds to the object's category. The bounding box transformation transforms an axis-aligned unit bounding box to the annotated bounding box in the world-coordinate system. For each frame in the video, we also compute the transformed bounding box in the camera coordinate system as well.

Besides raw data, we also provided a pre-processed dataset that can be easily connected to existing input pipelines for model training. We converted our entire dataset to Tensorflow tf.Example and tf.SequenceExample for image and video models, respectively. Both formats are stored as Tensorflow records. We implementing example pipelines to feed this data to PyTorch, Tensorflow, and Jax training pipelines efficiently. 

\section{Conclusion} \label{sec:conclusion}
This paper introduces the Objectron dataset: a large scale object-centric dataset of $14,819$ short videos in the wild with object pose annotation. 
We developed an efficient and scalable data collection and annotation framework based on on-device AR libraries. 
We also presented results of our proposed two-stage 3D object detection model trained on the Objectron dataset as the baseline. 
By releasing this dataset, we hope to enable the research community to push the limits of 3D object geometry understanding and foster new research and applications in 3D understanding, video models, object retrieval, view synthetics, and 3D reconstruction.


{\small
\bibliographystyle{ieee_fullname}
\bibliography{references}
}

\end{document}